%% file: main.tex
\documentclass{article}

\usepackage{microtype}
\usepackage{graphicx}
\usepackage{subfigure}
\usepackage{booktabs} 
\usepackage{multirow}
\usepackage[dvipsnames]{xcolor}

\usepackage{pdfpages}
\usepackage{enumitem} 
\usepackage{tcolorbox}
\usepackage[fixed]{fontawesome5}

\usepackage{listings}
\usepackage{xcolor}
\usepackage{subcaption}
\usepackage{tabularx}
\usepackage{amsmath} 

\usepackage[hyphens]{url}
\usepackage[hidelinks]{hyperref}
\hypersetup{breaklinks=true}

\usepackage[accepted]{icml2024}
\usepackage[symbol]{footmisc}

\icmltitlerunning{Project Alexandria: Towards Freeing Scientific Knowledge from Copyright Burdens via LLMs}

\begin{document}

\twocolumn[
\vspace*{-0.75cm}
\icmltitle{Project Alexandria: Towards Freeing Scientific Knowledge\\ from Copyright Burdens via LLMs}

\vspace*{-0.3cm}
\begin{center}
\textbf{ 
Christoph Schuhmann$^{1\circ}$ \quad
Gollam Rabby$^{3,1\circ}$ \\
Ameya Prabhu$^{2,4\diamond}$ \quad
Tawsif Ahmed$^{1,2\diamond}$ \quad
Andreas Hochlehnert$^{2,4\diamond}$ \quad
Huu Nguyen$^{1,2,5\diamond}$  \quad
Nick Akinci$^{6\diamond}$ \\
Ludwig Schmidt$^{2,7\dagger}$ \quad
Robert Kaczmarczyk$^{1,10\dagger}$ \quad
Sören Auer$^{3,8\dagger}$ \quad
Jenia Jitsev$^{1,2,9\dagger}$ \quad
Matthias Bethge$^{2,4\dagger}$ \vspace{0.1cm}}\\
{
$^1$LAION \quad
$^2$Open-$\Psi$ (Open-Sci) Collective \quad
$^3$L3S Research Center, Leibniz University Hannover \\ 
$^4$T\"ubingen AI Center, University of T\"ubingen \quad
$^5$Ontocord.ai \quad 
$^6$Heidrich Rechtsanwälte \quad
$^7$Stanford University \\
$^8$ TIB—Leibniz Information Centre for Science and Technology \quad 
$^9$ Juelich Supercomputing Center\\
$^{10}$ Technical University of Munich, Germany\\} \vspace{0.2cm}
\raisebox{-1pt}{\faDatabase} \href{https://huggingface.co/datasets/laion/project-alexandria}{\texttt{Knowledge-Units Database}} \quad 
\raisebox{-1pt}{\faGlobe} \href{https://projects.laion.ai/project-alexandria/}{\texttt{Project Page}} \quad 
\raisebox{-1pt}{\faGithub} \href{https://github.com/laion-ai/project-alexandria}{\texttt{Codebase}}
\end{center}
\vskip 0.3in
]
\footnotetext{\hspace{-0.4cm}$^\circ$ Equal lead \quad  $^\diamond$Equal core contribution \quad $^\dagger$Equal supervision\\ Refer to Author Contributions section for details.}

\input{sections/abstract}

\input{sections/intro}
\input{sections/knowledgeunits}
\input{sections/legal}

\input{sections/results}
\input{sections/alternate}
\input{sections/conclusions}

\bibliography{main}
\bibliographystyle{icml2024}

\newpage
\appendix
\input{supplementary/appendix}

\end{document}

%% file: sections/abstract.tex
\vspace*{-0.4cm}
\begin{abstract}
Paywalls, licenses and copyright rules often restrict the broad dissemination and reuse of scientific knowledge. We take the position that it is both \textit{legally} and \textit{technically} feasible to extract the scientific knowledge in scholarly texts.  Current methods, like text embeddings, fail to reliably preserve factual content, and simple paraphrasing may not be legally sound.
We propose a new idea for the community to adopt: convert scholarly documents into knowledge preserving, but style agnostic representations we term Knowledge Units  using LLMs. These units use structured data capturing entities, attributes and relationships without stylistic content. We provide evidence that Knowledge Units (1) form a legally defensible framework for sharing knowledge from copyrighted research texts, based on legal analyses of German copyright law and U.S. Fair Use doctrine, and (2) preserve most ($\sim$95\%) factual knowledge from original text, measured by MCQ performance on facts from the original copyrighted text across four research domains. 
Freeing scientific knowledge from copyright promises \textit{transformative benefits for scientific research and education} by allowing language models to reuse important facts from copyrighted text. To support this, we share open-source tools for converting research documents into Knowledge Units. Overall, our work posits the feasibility of democratizing access to scientific knowledge while respecting copyright.\vspace{-0.25cm}
\end{abstract}

%% file: sections/intro.tex
\section{Introduction}

Scientific publishing has grown tremendously in recent decades \cite{white2019publications}, but many researchers still lack access to crucial papers \cite{suber2012open}. This access gap exists even in the wealthiest academic libraries in the world and is much worse for Researchers in developing countries, small independent labs or for independent scholars and educators. In 2008, Harvard had 98,900 subscriptions, Yale 73,900 and the best-funded research library in India only 10,600 \cite{suber2012open}. This disparity, coupled with the unsustainable surge in academic journal subscription costs, has been a pivotal driving force behind the Open Access (OA) movement. Although OA has become more common, this \textit{scholarly communication crisis} still remains an issue today. In July 2018 up to 200 German institutions lost access to journals of Elsevier \cite{else2018dutch} for five years after unsuccessful negotiations. While the negotiations since succeeded this temporary cut from new publications had a measurable effect on publishing behavior \cite{NoDealFraser}. Furthermore, in 2019 major US universities canceled their subscriptions to certain journals stating unreasonable price surges as the reason \cite{gaind2019huge}. This barrier slows scientific progress and as such ``copyright laws work counter to prevailing scientific norms" \cite{stodden2008legal}. 

\textbf{Why this position paper?} Recent advances in language models (LLMs), like GPT-4 \cite{achiam2023gpt}, Llama \cite{touvron2023llama}, and recently Deepseek R1 \cite{deepseekai2025deepseekr1incentivizingreasoningcapability} allow us to democratize access to scholarly knowledge, such as answering questions based on existing scientific research. The core issue lies in scholarly texts containing both \textit{information} and \textit{artistic} elements — such as wording, style, and unique phrasing — of which the latter is protected by copyright. With LLMs, for the first time, we can extract knowledge at scale and free scientific content while respecting authors' rights to copyright.

\textbf{Contribution.} In this paper, we take the position that separating the factual information in scholarly works from the copyrighted creative expression is \textit{technically and legally feasible}. We advocate for a \textit{Project Alexandria}, 
to realize this vision by creating \textit{Knowledge Units} using LLMs, which systematically separate reusable \textit{information} from the \textit{artistic expressions} inherent in scholarly writings. \textit{Knowledge Units} are structured records capturing entities, relationships, and attributes extracted from scholarly texts in a database.
Our study tests the \textit{legal} and \textit{technical} feasibility of preserving scientific knowledge that researchers and students need to learn or apply. We posit it must satisfy two desiderata:

\textit{\textbf{1. Legal Defensibility.}} Under interpretations of German copyright law and the U.S. Fair Use doctrine, facts themselves are not subject to copyright protection, only their creative expression. In addition to these jurisdictions, concept of the idea-expression dichotomy is also adopted by many other jurisdictions, such as the UK and India \cite{jain2012principle, adhikari2021idea}. We provide a legal analysis showcasing that Knowledge Units could preserve only the factual substance — definitions, measurements, causal relationships, and methodological details. They are designed explicitly to exclude the original phrasing and stylistic elements, offering what we believe is a legally permissible way to share scientific knowledge openly.

\textit{\textbf{2. Information Preservation.}} We evaluate the fidelity of Knowledge Units through performing question-answering (QA) experiments using them, where language models answer multiple-choice questions based on abstracts and full-text articles across biology, physics, mathematics, and computer science. Models provided with Knowledge Units achieve high accuracy, nearly matching the performance of those given the original texts. This provides evidence that the vast majority of relevant information should be accurately captured in this format.

\textbf{Impact.} Freeing factual information from copyright restrictions can help more researchers, educators, and language models access and share facts from scholarly work.  As a result, researchers worldwide can discuss and build on one another’s findings without legal uncertainty or relying on expensive paywalls. We believe this is a key step toward a more open and inclusive global research community. We present a clear vision, a practical mechanism, and an open-source infrastructure aimed at fostering a more inclusive and collaborative global scientific ecosystem.

In the rest of this paper, we present how Knowledge Units work, discuss the legal principles that support this approach, show evidence of high factual retention, and consider alternate perspectives. We conclude by describing future plans to build massive open databases of factual knowledge and encourage the community to adopt this approach.

%% file: sections/knowledgeunits.tex
\vspace{-0.2cm}
\section{Knowledge Units}

To extract knowledge from a document, its text is segmented into manageable sections or paragraphs. Each paragraph is subsequently processed by an LM to identify entities, along with their attributes and relationships, resulting in a structured output referred to as a \textit{Knowledge Unit} (Table \ref{tab:original_vs_ku}).

\vspace{-1.0em}

\begin{quote}
\textbf{Knowledge Unit (KU):} A set of entities, attributes, and relationships, capturing a short original text excerpt. 
\end{quote}

\vspace{-1.0em}
Each Knowledge Unit captures:
\begin{itemize}[leftmargin=*,noitemsep,topsep=0em]
    \item \textit{Entities}: the core concepts or objects in the paragraph, with relevant attributes.
    \item \textit{Relationships}: statements that connect or link entities, such as causal or definitional relationships.
    \item \textit{Attributes}: statements that describe entities according to the excerpt.
    \item \textit{Context summary}: A few sentences summarizing the previous knowledge units.
    \item \textit{Sentence MinHash}: A list of MinHashes of the source sentences used to generate this KU. 
\end{itemize}

\begin{table*}[ht]
\centering
\small
\vspace{-0.3cm}
\caption{\textbf{Converting Original Text to a Knowledge Unit.} \emph{Left:} Sample input paragraph presenting the dark matter field fluid model explaining the Earth-Moon system's evolution and predicting Mars's rotational deceleration, challenging tidal friction as the primary driver.~\cite{pan2007evolution} 
\emph{Right:} Resulting \textbf{Knowledge Unit}, including a short contextual note, MinHashes of the source sentences and extracted factual statements (entities, relationships, attributes). The output has been truncated, a complete KU can be found in the Appendix \ref{sec:example_ku}.}\vspace{0.1cm}
\setlength{\tabcolsep}{8pt} 
\renewcommand{\arraystretch}{1.3} 
\begin{tabularx}{\textwidth}{X|X}
\hline
\textbf{Original Text} & \textbf{Knowledge Unit Representation} \\
\hline
The evolution of the Earth-Moon system is described by the dark matter field fluid model proposed in the Meeting of Division of Particle and Field 2004, American Physical Society. The current behavior of the Earth-Moon system agrees with this model very well, and the general pattern of the evolution of the Moon-Earth system described by this model agrees with geological and fossil evidence. The closest distance of the Moon to Earth was about 259000 km at 4.5 billion years ago, which is far beyond the Roche's limit. The result suggests that the tidal friction may not be the primary cause for the evolution of the Earth-Moon system. The average dark matter field fluid constant derived from Earth-Moon system data is $4.39 \times 10^{-22} \, \text{s}^{-1}\text{m}^{-1}$. This model predicts that Mars's rotation is also slowing with the angular acceleration rate about $-4.38 \times 10^{-22} \, \text{rad} \, \text{s}^{-2}$. 
& 
\begin{minipage}[t]{\linewidth}
\tiny
\textbf{Context:} \texttt{"The text appears in a scientific discussion on how the Earth-Moon system's evolution can be explained by a dark matter field fluid model\ldots"}\\
\textbf{Source Sentence MinHash:} \texttt{[24175356, 47043276, 9024081, 8553571, \ldots]},\\
\textbf{Earth-Moon System}: \texttt{\{} \\
    \hspace*{0.5em} \textbf{Relations}: \texttt{\{} \\
        \hspace*{1em} \texttt{evolution\_described\_by: "Dark Matter Field Fluid Model"},\\
        \hspace*{1em} \texttt{current\_behavior\_agrees\_with}: \texttt{"Dark Matter Field Fluid Model"},\\
        \hspace*{1em} \texttt{evolution\_pattern\_agrees\_with}: \texttt{["Geological Evidence", "Fossil Evidence"]} \\
    \hspace*{0.5em} \texttt{\}}, \\
    \hspace*{0.5em} \textbf{Attributes}: \texttt{\{} \\
        \hspace*{1em} \texttt{closest\_distance\_4.5\_billion\_years\_ago}: \texttt{"259000 km"},\\
        \hspace*{1em} \texttt{distance\_relative\_to\_Roche\_limit}: \texttt{"Far beyond"} \\
    \hspace*{0.5em} \texttt{\}} \\
\texttt{\},}\\
\textbf{Dark Matter Field Fluid Model}: \texttt{\{} \\
    \hspace*{0.5em} \textbf{Relations}: \texttt{\{} \\
        \hspace*{1em} \texttt{proposed\_at}: \texttt{"Meeting of Division of Particle and Field 2004, American Physical Society"},\\
        \hspace*{1em} \texttt{describes\_evolution\_of}: \texttt{"Earth-Moon System"},\\
        \hspace*{1em} \texttt{predicts\_slowing\_rotation\_of}: \texttt{"Mars"} \\
    \hspace*{0.5em} \texttt{\}}, \\
\ldots \\
\end{minipage}
\\
\hline
\end{tabularx}
\vspace*{-0.5cm}
\label{tab:original_vs_ku}
\end{table*}

\textbf{Motivation for Paragraph-Level Granularity}
Using paragraph-sized segments avoids two extremes. If segments are too short (such as single sentences), the knowledge becomes scattered. If segments are too long (such as entire papers), the language model's question-answering performance deteriorates, as it struggles to effectively extract all relevant facts from such large chunks of text in a single step. Paragraph-level chunks represent an optimal granularity that allows the language model to focus effectively on extracting and processing all the facts contained within the text segment.

\textbf{Comparison to Knowledge Graphs}
Knowledge Units are very similar to knowledge graphs~\cite{hogan2021knowledge}; however, they are generated locally, on a per-paragraph basis, rather than globally. One primary difference is that Knowledge Graphs require to have either locally or even globally unique identifiers (URI/IRIs).  This is not the case for the KU graphlets we extract - for two KUs extracted from neighboring paragraphs of a source document, the entities and relationships might differ, even when referring to the same concepts. Hence, KUs lack more logic-based knowledge representation such as OWL-based knowledge graphs. Secondly, unlike conventional knowledge graph tools such as REBEL~\citep{rebel2021}, which decompose text into concise triples (subject, relation, object), Knowledge Units maintain the contextual richness and nuances of the original paragraph, ensuring that relationships and entities remain closely aligned with the source text. While these are significant disadvantages for general-purpose knowledge representation, it is not required for our prime use case of knowledge and context provision for downstream AI applications.

\textbf{Constructing Knowledge Units with LLMs}
\label{sec:ch3:ku_pipeline}
We create each Knowledge Unit using large language models guided by a few-shot prompt. We split each scholarly text into paragraphs that have roughly 200 to 500 tokens. The language model then extracts a structured set of entities, relationships, and attributes from each paragraph. We instruct the model to avoid copying the original wording. Instead, it stores key facts in a simple data structure, omitting stylistic language. If we process a longer document, we include the previous 10 KUs in the prompt so that the model remains consistent in naming entities across paragraphs.

%% file: sections/legal.tex
\vspace{-0.2cm}
\section{Legal Considerations for Fair Knowledge Extraction}

We start our legal considerations under the broader human rights framework of the Universal Declaration of Human Rights (UDHR), Article 27(a) which states ``[e]veryone has the right freely to participate in the cultural life of the community, to enjoy the arts and to share in scientific advancement and its benefits." However, it also specifies the need to support and respect the rights of authors in their “moral and material interests resulting from any scientific, literary or artistic production of which (s)he is the author” as stated under the UDHR 27(b). These rights of authors must co-exist with everyone's rights to scientific knowledge, and indeed freeing such knowledge from copyright burdens will allow scholars to participate in the marketplace of ideas - to continue new research. From this starting point, we analyzed in detail the legal implications of extracting knowledge from copyrighted text under two legal frameworks: German and U.S. copyright law, presented below.

\subsection{German Copyright Law}

Under German law, \textbf{Urheberrecht} (copyright) serves to protect creative expressions, ensuring that authors maintain exclusive rights over their original works \cite{wandtke2010urheberrecht, geller2009german}. Therefore, only the aesthetic design of a text is protected, but not the content itself. An important exception exists for complex narratives that stem from an author's imagination, which may receive copyright protection. In contrast, mere facts and scientific discoveries remain unprotected.

For a work to qualify for copyright protection, it must satisfy the \emph{Schöpfungshöhe} (original creative threshold) \cite{wandtke2010urheberrecht, hoffmann2011copyrights}, and only human authors — excluding AI systems — can hold authorship rights \cite{gratz2021kunstliche, legner2019erzeugnisse}. The \emph{extraction of information} from copyrighted texts does not infringe upon the original rights holder’s exclusive privileges \citep{bmbf2020copyright}, provided three specific conditions are met: no protected text is copied, the extracted text is fact-centric, and data mining exemptions are adhered to.

\textbf{Accessing Original Text:} According to Sections 44b and 60d of the \emph{Urheberrechtsgesetz} (UrhG), or the German Copyright Act, there is explicit permission to temporarily store copyrighted works for the purpose of extracting insights such as patterns, trends, or correlations, particularly in the context of scientific research (\emph{Text and Data Mining} )~\citep{heidrich2023legal}. Both provisions are exceptions to the principle that works protected by copyright may only be reproduced with the permission of the author or rights holder. Organizations operating as non-profits in the scientific research sector that engage in large-scale text analysis are offered robust support under \S44b and \S60d UrhG, being authorized to perform such activities, provided they systematically delete the underlying works once the factual extraction process is complete.

\textbf{Protected Text is Not Extracted:} The process of abstracting or summarizing the main ideas from a work can be permissible under German copyright law, \emph{provided the summary does not replicate the original’s creative form} \citep{heidrich2023legal}. Knowledge Units aim to comply to this principle by avoiding the storage of even summary phrases. Instead, it focuses solely on maintaining relationships, domain-specific concepts, or numeric attributes. If Knowledge Units do not reveal the text’s distinctive arrangement or style, they cannot be deemed “unfreie Bearbeitung.” (unfree adaptation) 

\textbf{Publication and Use of the Extracted Knowledge Units:} Under \S15 UrhG, authors retain control over the reproduction and public communication of their works. However, when it comes to the publication of extracted Knowledge Units, if what is being disseminated is \emph{non-protected factual content} and the creative aspects of the original work are \emph{not} reproduced in this process, they would be exempt. We highlight that any new textual or data-based work that arises solely from \emph{factual extraction} is inherently authored by those who develop the new structure. Such works may be considered unprotected if they are exclusively machine-generated without any human creative input, aligning with the legal framework that restricts authorship rights to human creators~\cite{heidrich2023legal}. As a result, it is viable to publish these Knowledge Units without infringing upon the original author’s rights. 

Overall, the processes of accessing original texts, extracting Knowledge Units, and publishing these extracted units can be conducted in a manner that neither reproduces nor stores the original phrasing, sentence structures, or distinctive literary qualities of the source material. This allows that there is no \emph{unfreie Bearbeitung} (unfree adaptation) involved, thereby avoiding any breach of German copyright laws. By explicitly prompting, we seek to adhere to the stipulated conditions — avoiding the copying of protected text, focusing on fact-centric extraction, and following data mining exemptions — allowing the  extracted Knowledge Units to be published while complying with the legal protections afforded to original creative works under German law.

\subsection{The Idea-Expression Dichotomy and US Fair Use Doctrine in Copyright}

In the United States, the legal framework regarding copyright differs slightly from Germany but leads to similar conclusions as other jurisdictions. U.S.\ copyright law does not protect facts or ideas, only the \emph{expressions} of those facts, an axiom that courts have repeatedly affirmed. Notably, the \textbf{Fair Use} doctrine (17 U.S.C.\ \S107) also allows provides a flexible framework that accommodates new and transformative uses such as text and data mining (TDM)~\citep{USfairuseTDM2015, reichman2012copyright}. The basis for using the knowledge downstream is the idea-expression dichotomy \citep{yen1989first} codified in 17 U.S.C.\ \S102 (b):  
\vspace{-1.1em}
\begin{quote}
    In no case does copyright protection for an original work of authorship extend to any idea, procedure, process, system, method of operation, concept, principle, or discovery, regardless of the form in which it is described, explained, illustrated, or embodied in such work.
\end{quote}
\vspace{-0.5em}
Title 17, Section 107 of the U.S.\ Code also enumerates four factors to evaluate fair use. For TDM, the first factor, \textbf{Purpose/Character}, demands for output text (in our case, Knowledge Units) to be typically “highly transformative,” especially if the copying is for nonprofit research or distinct from the original text’s use. The second factor, \textbf{Nature of work}, although many TDM cases involve creative works, courts have often downplayed or treated this factor as neutral if the use remains transformative. The third factor, \textbf{Amount/Substantiality}, permits copying \emph{the entire text} to achieve meaningful analysis, with courts finding that the “all or nothing” nature of TDM demands full copying~\citep{USfairuseTDM2015}. Lastly, the fourth factor, \textbf{Effect on market}, is generally favorable for TDM because it does not serve as a substitute for reading or consuming the original work, thereby rarely damaging the market for the original.

When these factors are weighed collectively, courts usually find that TDM constitutes fair use—particularly in academic or research settings~\citep{USfairuseTDM2015, hathitrustcase}. Knowledge Units aim to operate in accordance with the best practice recommendations from “The Code of Best Practices in Fair Use for Academic and Research Libraries,” which explicitly endorses the creation of TDM databases, provided that full-text or near-verbatim distributions are not made publicly available~\citep{USfairuseTDM2015}. Our pipeline aims to adheres to these recommendations by ensuring that no substantial original expression is published; Knowledge Units contain only factual statements, short style descriptors, and minimal numeric references. Additionally, there is no end-user access to entire works, as the original copyrighted text is neither exposed nor distributed. Instead, it is either deleted after analysis or stored for ephemeral TDM tasks within the scope of allowable research usage. Furthermore, we focus on non-consumptive research, providing derived knowledge for advanced AI, searching, and research purposes, rather than for reading or substituting the original content.

\textit{Past Relevant Case Law} illustrates the application of these principles. In \emph{Authors Guild v. HathiTrust}, the court emphasized that scanning entire works to facilitate full-text search and enable computational analysis was \emph{“quintessentially transformative”} \cite{hathitrustcase}. Similarly, in \emph{Authors Guild v. Google}~\cite{campbell2016authors}, the massive digitization for Google Books was held to be fair use, partly because it “transformed the book text into data for the purpose of substantive research” \cite{googleBooksCase}.

Overall, our approach likely aligns with past cases and could gain broad acceptance under U.S.\ fair use precedents, as it fosters \textit{public-interest scholarship} with measures taken to avoid threatening authors’ legitimate markets or moral rights to their expression. In the U.S. context, such re-purposing contributes distinct value and fosters new lines of inquiry, such as large-scale pattern identification. Moreover, it does not replace the original text as reading material, minimizing risk of market harm.

%% file: sections/results.tex
\vspace{-0.2cm}
\section{Evaluating Knowledge Unit Effectiveness}

\begin{table*}[ht]
\centering
\small
\vspace*{-0.25cm}
\caption{\textbf{Knowledge Unit Performance Across Domains (Abstract-Level Analysis).} Each column displays the lower-upper bound performance (no context vs. original text) and the Knowledge Unit (KU) performance for different models. Using KUs preserves most information for answering MCQs, perform close to the using the original text (upper bound) across domains and models.}\vspace{0.1cm}
\setlength{\tabcolsep}{4pt}
 \resizebox{0.9\textwidth}{!}{
\begin{tabular}{lcccccccc}
\toprule
\textbf{Model} &
\multicolumn{2}{c}{\textbf{Medical}} &
\multicolumn{2}{c}{\textbf{Computer Science}} &
\multicolumn{2}{c}{\textbf{Mathematics}} &
\multicolumn{2}{c}{\textbf{Physics}} \\
\cmidrule(lr){2-3} \cmidrule(lr){4-5} \cmidrule(lr){6-7} \cmidrule(lr){8-9}
& \textbf{[Lower-Upper]} & \textbf{KU} & \textbf{[Lower-Upper]} & \textbf{KU} & \textbf{[Lower-Upper]} & \textbf{KU} & \textbf{[Lower-Upper]} & \textbf{KU} \\
\midrule
Gemini (1.5-Flash 002) & 42.28--97.17 & 93.37 & 58.56--97.27 & 93.62 & 52.26--94.68 & 91.82 & 34.19--95.30 & 92.97 \\
Qwen 2.5 (7B) & 42.76--97.00 & 92.76 & 58.59--97.27 & 93.45 & 52.29--94.79 & 92.87 & 36.80--95.29 & 92.97 \\
Mistral Small (Dense 22B) & 42.46--97.13 & 92.33 & 58.79--97.37 & 94.70 & 52.33--94.74 & 92.91 & 34.58--95.38 & 90.56 \\
Ministral 2410 (3B) & 42.24--97.06 & 88.22 & 58.48--97.36 & 91.91 & 52.21--94.80 & 87.65 & 33.03--95.29 & 87.14 \\
Llama 3.2 (3B) & 42.52--97.10 & 87.08 & 58.63--97.34 & 88.47 & 51.61--94.82 & 86.44 & 36.89--95.33 & 86.90 \\
Llama 3.1 (8B) & 42.69--97.13 & 85.80 & 58.68--97.31 & 87.75 & 52.00--94.81 & 84.21 & 37.04--95.29 & 85.43 \\
\bottomrule
\end{tabular}}
\vspace*{-0.4cm}

\label{tab:performance_models_vertical}
\end{table*}

\begin{table}[ht]
    \centering
    \small
    \vspace{-0.25cm}
    \caption{\textbf{Knowledge Unit Performance in Longer Documents.} Multiple-choice performance with knowledge units (KU) remains far above the no-context baseline and approaches the original-text upper bound, though it is slightly lower for long documents. This indicates that using KU context preserves most information across different domains and models, with small degradation.}\vspace{0.1cm}
    \setlength{\tabcolsep}{3pt}
    \resizebox{\linewidth}{!}{
    \begin{tabular}{lcccc}
        \toprule
        \textbf{Model} & 
        \multicolumn{2}{c}{\textbf{Physics}} & 
        \multicolumn{2}{c}{\textbf{Medical}} \\
        \cmidrule(lr){2-3} \cmidrule(lr){4-5}
        & \textbf{[Lower-Upper]} & \textbf{KU} & \textbf{[Lower-Upper]} & \textbf{KU} \\
        \midrule
        Gemini (1.5-Flash 002) & 49.48--90.72 & 83.51 & 46.96--94.13 & 81.76 \\
        Qwen 2.5 (7B) & 52.23--89.69 & 79.04 & 50.45--93.24 & 88.29 \\
        Mistral Small (Dense 22B) & 50.86--89.35 & 81.44 & 48.31--94.59 & 90.20 \\
        \bottomrule
    \end{tabular}}
     \vspace{-0.35cm}
    \label{tab:performance_models_extended_vertical}

    \vspace{-0.35cm}
\end{table}

One major question is whether removing creative expression still preserves enough factual information to be useful. We designed multiple experiments to study these issues.

\vspace{-0.2cm}
\subsection{Experimental Setup and Design}
\label{sec:exp_setup}

\textbf{Design Principles.} Our evaluation requires scalable benchmarks that adapt across different research domains. To achieve this, we adopt a multiple-choice question (MCQ) design with questions, correct answers and distractors generated by a frontier LM. We chose MCQs for three key reasons: (1) They confirm with users querying knowledge contained in specified original text, providing clearer insights than retrieval metrics; (2) Not requiring manual generation allows for easy customization across various disciplines; (3) Repeating benchmark generation and retesting helps capture broader variance and reduces the risk of overfitting.

\textbf{Key Idea.} We tested how well multiple-choice question (MCQ) performance is preserved when we convert the original text into Knowledge Units data. We created MCQs for each text excerpt, asked language models to answer them with no context (lower bound), then asked them again with the original text (upper bound) for sanity check since they are automatically generated. Finally, we tested them with only the Knowledge Units (our method).

\textbf{Datasets.} We used: 
\begin{enumerate}[leftmargin=*,noitemsep,topsep=0em]
    \item \textit{Abstract-level analysis}: 1{,}000 abstracts each from Biology, Mathematics, and Physics of the peS2o dataset~\cite{peS2o} as well as 1{,}000 abstracts from Computer Science from ArXiv~\cite{clement2019arxiv}. For each abstract, we generate three MCQs (Appendix, Table \ref{tab:abstract_medical_mcqs},~\ref{tab:abstract_math_mcqs},~\ref{tab:abstract_cs_mcqs},~\ref{tab:abstract_phy_mcqs}) and one KU.
    \item \textit{Full-paper analysis}: 200 longer papers (100 Medical, 100 Physics)~\cite{DBLP:conf/naacl/CohanDKBKCG18}. We chunked each paper into segments of 200 words and generated Knowledge Units for each chunk, referencing the previous 10 units for context continuity. For each paper we generate 10 MCQs (Appendix, Table \ref{tab:fullpaper_medical_mcqs} and~\ref{tab:fullpaper_phy_mcqs}).
\end{enumerate}

\textbf{Procedure:} The Gemini Pro 1.5 002 model was utilized to generate all MCQs based on the original text, as well as to provide annotations for the correct answer. The questions are designed to assess specific, verifiable elements, such as factual claims, numerical data, definitions, or relational knowledge, to minimize ambiguity. The accuracy of the answers is verified through cloze evaluation, as is standard in the LM-Harness.

\begin{table*}[ht]
    \centering
    \small
    \vspace*{-0.25cm}
    \caption{\textbf{Average and Top 5\% n-gram Overlap and Plagiarism Check Scores for Gemini (1.5-Flash) and Qwen 2.5 (7B).} n-gram overlap measures Jaccard similarity for sequences of n words. Plagiarism scores indicate textual similarity, with values below 20\% considered negligible. Both metrics remain low between Input Texts and Knowledge Unit or corresponding reconstructed text, preliminarily suggesting minimal direct reproduction of the original text in Knowledge Units.
    } \vspace{0.1cm}
\setlength{\tabcolsep}{4pt}
 \resizebox{0.7\linewidth}{!}{
\begin{tabular}{llcccc}
    \toprule
    \textbf{Model} & \textbf{Data} & \textbf{Plagiarism Score} & \textbf{5-gram Overlap} & \textbf{7-gram Overlap} & \textbf{11-gram Overlap} \\
    \midrule
    \multicolumn{6}{c}{Original Text and Knowledge Unit Overlap}\\ \midrule
    \multirow{2}{*}{Gemini-1.5 Flash} & Overall & 2.7 & 0.009 & 0.003 & 0.001 \\
     & Top 5\% & 14.5 & 0.023 & 0.011 & 0.003 \\ \cmidrule{2-6}
    \multirow{2}{*}{Qwen-2.5 (7B)}           & Overall & 5.9 & 0.028 & 0.015 & 0.005 \\
             & Top 5\% & 22.9 & 0.070 & 0.047 & 0.024 \\ \midrule
    \multicolumn{6}{c}{Original Text and Reconstructed Text Overlap}\\ \midrule
   
\multirow{2}{*}{Gemini-1.5 Flash} 
 & Overall & 3.8  & 0.022 & 0.010 & 0.002 \\
 & Top 5\% & 12.1 & 0.047 & 0.030 & 0.013 \\\cmidrule{2-6}
  \multirow{2}{*}{Qwen-2.5 (7B)} 
 & Overall & 17.8 & 0.142 & 0.123 & 0.098 \\
 & Top 5\% & 24.0 & 0.175 & 0.157 & 0.133 \\

    \bottomrule
\end{tabular}}
\vspace*{-0.4cm}
\label{tab:average_sherlock_ngram_scores}
\end{table*}

\vspace{-0.2cm}
\subsection{Results: Information Retention}

We evaluate the effectiveness of Knowledge Units (KUs) in preserving information from original text by addressing two key questions:

\textbf{Q1: Does conversion retain information from the original text chunks (Abstract-Level Analysis)?}  To answer this, we compared multiple-choice question (MCQ) performance using Knowledge Units against performance using the original text. We tested several small language models, first asking them to answer MCQs based on full original passages, then repeating the test with only Knowledge Units. Table~\ref{tab:performance_models_vertical} presents the results.

\textit{Findings.} As expected, models answering without any context (lower bound) performed significantly worse than those given the full original text (upper bound), confirming that context is crucial for correctly answering the MCQs. While language models could occasionally eliminate incorrect answer choices using prior knowledge, their accuracy remained low without context. When provided with Knowledge Units instead of the original text, model performance closely matched the upper bound in nearly all cases. This suggests that Knowledge Units preserve the majority of relevant information needed for answering questions. The pattern held across different models and research domains. Additionally, the variance in performance across different question sets remained between 3–5\%, indicating a statistically significant — but relatively small — difference between using the original text and using Knowledge Units.

\textbf{Q2: Can conversion preserve information in long documents (full paper analysis)?}  To examine this, we assessed model performance on long-document MCQs, where each document was segmented into multiple Knowledge Units. We used models with larger context windows, both to generate the Knowledge Units and to answer the questions. Table~\ref{tab:performance_models_extended_vertical} summarizes the results.

\textit{Findings.} Consistent with Q1, the gap between the lower and upper bounds remained large, reaffirming the validity of our MCQ evaluation approach. However, we observed a slight decline in the upper bound performance, suggesting that longer documents introduced additional complexity, making some questions harder to answer even with access to the full text. Knowledge Units performed slightly worse in long-document scenarios compared to short text segments. This suggests that aggregating multiple Knowledge Units across long contexts introduces some challenges in reasoning. However, performance remained much closer to the upper bound than the lower bound, confirming that Knowledge Units still retained most of the critical information. As with Q1, variance across different question sets remained within 3–5\%, reinforcing the reliability of these findings.

\textbf{Conclusion.} Knowledge Units effectively preserve information across multiple domains. While performance degradation is observed in long-document scenarios, the majority of factual content remains intact. These results are consistent across different model families and scientific disciplines, demonstrating the robustness of Knowledge Units as a structured knowledge representation format.

\subsection{Results: Assessing Content Overlap}

We have established that key information is preserved in the generated Knowledge Units (KUs). As a next step, we perform empirical checks to detect potential text reuse. In legal contexts, consistent high n-gram overlap across large spans of text is a commonly used measure to prove textual reuse, though not a definitive measure by any means. Here, we compute n-gram overlaps between the original abstracts and our generated KUs on an abstract-level dataset, with results shown in Table~\ref{tab:average_sherlock_ngram_scores} for the two best models.

\textbf{Q1: Is there a significant overlap?} The top portion of Table~\ref{tab:average_sherlock_ngram_scores} reports 5-gram, 7-gram, and 11-gram Jaccard similarities for the entire dataset, as well as for the top 5\% of the most similar original text–KU pairs. The Gemini-1.5 Flash model consistently exhibits very low overlap ($<$3\% in the most conservative 5-gram scenario), even in the top 5\% subset. The Qwen-2.5 model shows slightly higher scores but remains below 7\% in the same scenario. Overall, these findings indicate negligible direct textual reuse.

\textbf{Q2: Does a plagiarism check show different trends?} We additionally use an open-source plagiarism detector~\cite{sherlock}, which attempts to detect more subtle forms of reuse (e.g., paraphrasing, synonym substitutions). Scores below 20\% are typically dismissed as negligible threshold and are not even displayed, according to official documentation. We apply this check instead of the n-gram metric, keeping everything else constant.  Table~\ref{tab:average_sherlock_ngram_scores} reports averages of 3--5\% for the entire dataset, rising to 15--23\% in the top 5\% of most similar pairs. These numbers remain below or close to even automatic dismissal thresholds, let alone a conservative actionable plagiarism check. This reinforces the findings from n-gram scores, indicating no direct text reuse.

\textbf{Q3: Does reconstructing text increase overlaps?} We next tried our best prompting a strong LM (Gemini-1.5 Pro) to regenerate the original abstract from a knowledge graph, using few-shot examples from the same domain. While Qwen-2.5 shows a substantial increase in n-gram overlap in the top 5\% subset, overlaps for both models, especially Gemini-1.5 Flash, remain extremely low in an absolute sense. Sherlock scores also remain unchanged in all but overall the Qwen-2.5 case. Manual inspection of the highest-overlap (11-gram) passages, with examples provided in Appendix  \ref{sec:appsimilarity}, suggests that most identical segments are filler phrases common in scientific writing, rather than substantive stylistic overlap.

\textbf{Conclusion.} Across both n-gram overlap measures and a dedicated plagiarism detector, evidence of direct text copying remains minimal. Even when models are explicitly prompted to reconstruct the source text, stylistic carryover is surprisingly low. While neither n-gram overlap nor plagiarism checks are legal standards for copyright, they provide preliminary empirical reassurance that the original text is not being substantially reproduced. 

%% file: sections/alternate.tex
\vspace{-0.2cm}
\section{Alternative Positions}
\vspace{-0.1cm}

Extracting valuable scientific knowledge from scholarly texts is debated from two main perspectives. One view argues that existing methods like LM embeddings already separate factual content from expressive elements, making a new format like Knowledge Units unnecessary (see Section \ref{sec:embed}). Conversely, critics question whether scientific knowledge can ever be freed from copyright constraints, citing the complexity of legal challenges, limitations of automated extraction methods, and the potential for large-scale harm (see Subsection \ref{sec:criticism}).

\vspace{-0.2cm}
\subsection{Limitations of Embeddings}
\label{sec:embed}

\begin{table}[t]
\centering
\small
\vspace{-0.25cm}
\setlength{\tabcolsep}{6pt} 
\caption{Cosine similarity scores between original texts and their modified versions using BGE-M3 embeddings. Scrambled word orders achieve high similarity scores highlighting the embeddings' coarse-grained semantic representations.}\vspace{0.1cm}\label{tab:cosine_similarity}
\begin{tabular}{lc}
\toprule
\textbf{Texts Compared} & \textbf{Cosine Similarity} \\ \midrule
Original-Random Words (\textit{Lower Bound}) & 0.45 \\
Original-Unrelated Abstract & 0.47 \\ 
Original-Knowledge Unit  & 0.82 \\ 
Original-Scrambled Word Order & 0.89 \\
Original-Original (\textit{Upper Bound}) & 1.0\\ \bottomrule
\end{tabular}
\vspace{-0.5cm}
\end{table}

While text embeddings are commonly used to store and share even copyrighted content, we demonstrate they inadequately preserve scientific knowledge—even state-of-the-art models on MTEB Leaderboard \cite{muennighoff2023mtebmassivetextembedding} like BGE-M3 \cite{chen2024bgem3embeddingmultilingualmultifunctionality}. Embeddings primarily capture coarse semantic similarity but fail to encode precise factual statements, causal relationships, or numeric details.

\textbf{Experimental Setup:} We embedded abstracts from prior analysis using BGE-M3 and evaluated knowledge retention via cosine similarity. 

\textbf{Sanity Checks:} Baselines included (1) gibberish vs. original (lower bound), (2) original vs. itself (upper bound), (3) unrelated domain-matched abstracts, and (4) scrambled abstracts (randomized word order).

\textbf{Result.} Table~\ref{tab:cosine_similarity} summarizes the results. We see that heavily scrambled text often showed high similarity to the original, revealing that surface-level spurious patterns drive much of the model’s similarity score. Cosine similarity, the most popular method for using embeddings, cannot separate whether the extracted facts (e.g., relationships, causal statements, numerical data) actually match the source.

\textbf{Conclusion.} Embeddings often fail to capture precise factual details, making them unreliable \textit{technically} for preserving scientific knowledge. Similarly, simple paraphrasing may still resemble the original text’s structure and style too closely, raising potential \textit{legality} concerns.

\vspace{-0.2cm}
\subsection{Addressing Common Criticisms}
\label{sec:criticism}

We address a few common criticisms to our position below:

\textbf{1. Credit Attribution:} Critics contend that open-access extraction of research findings into structured databases risks diluting traditional citation metrics (e.g., impact factors), as users may cite the database over original papers.\\
\textit{Rebuttal:} Traceable attribution systems (e.g., DOIs embedded in extracted facts) and enhanced accessibility can amplify citation reach while preserving credit to authors.

\textbf{2. Oversimplification of Nuance in Research:} Knowledge Units may not effectively capture intricate procedures such as mathematical proofs or biochemical assays. In general, information that is best presented through tables, diagrams, or proofs does not easily convert into Knowledge Units.\\
\textit{Rebuttal:} We agree, this is a critical limitation of KUs. However, mathematical proofs, assay descriptions, tables of results and similar long-form structured texts are largely not even eligible for copyright. However, we need to accurately identify and release such text information as-is.

\textbf{3. Legal Risks:} Transformative-use defenses under fair use law are context-dependent, requiring costly case-by-case litigation, creating deterrence through legal uncertainty.\\
\textit{Rebuttal:} We agree. We try to mitigate this by careful design, using structured extraction frameworks converting text into non-expressive factual units. We additionally highlight the process must involve only temporary access to original copyrighted texts, strictly for the purpose of analysis under TDM exceptions, followed by their deletion. No original copyrighted material should be retained. 

\textbf{4. Hallucination Propagation:}  Automated extraction risks embedding inaccuracies, particularly in high-stakes domains, without scalable human validation.\\
\textit{Rebuttal:} We agree, and believe a critical next step is designing hybrid systems integrating cross-referencing algorithms, confidence scoring, and targeted human oversight for critical assertions to balance scalability with rigor. The KUs would be akin to editing a machine-readable Wikipedia (human-readable with good UI), one can simply edit the incorrect KU and replace the wrong information. Moreover, we believe that upcoming more capable models, like recently developed reasoning models, will provide further way to automate and scale verification of extracted knowledge content, requiring human assistance only in few manageable cases.

\textbf{5. Irreversible Harm.} Once Knowledge Units is released, any flaws discovered later affects all released content, making the copyright harm irreversible.\\
\textit{Rebuttal:} Our proposal presents an idea with a prototype to demonstrate feasibility -- this is a position paper rather than a proposed system. We stress the need for further iteration before deployment and advocate for exploration over immediate adoption to prevent uncritical implementation. Other measures, designed to reduce hallucination propagation can similarly update KU contents in this case, mitigating any harm caused similar to those adopted for leaks of copyrighted data.

\textbf{6. Replication of failures of Semantic Web.} We think semantic web initiatives such as schema.org or life-science KGs have excelled in adoption and impact. However, the standardization done in fully fledged KGs is not (yet) feasible (e.g., wrt. unique identification and schema harmonization) and hence omitted in the design of KUs to avoid past failures, as it is not critical for our purpose. We want to enable KUs to primarily allow sharing of knowledge on platforms like HuggingFace contribute strongly to open science.

%% file: sections/conclusions.tex
\vspace{-0.2cm}
\section{Impact: Why Free Scientific Knowledge?}
\vspace{-0.1cm}

Historically, making knowledge widely available has driven transformative progress. Gutenberg’s printing press broke medieval monopolies on information, increasing literacy and contributing to the Renaissance and Scientific Revolution. In today's world, open source projects such as GNU/Linux and Wikipedia show that freely accessible and modifiable knowledge fosters innovation while ensuring creators are credited through copyleft licenses. These examples highlight a key idea: \textit{access to essential knowledge supports overall advancement.} 

This aligns with the arguments made by Prabhakaran et al. \cite{humanrightsbasedapproachresponsible}, who specifically highlight the \textbf{ human right to participate in scientific advancement} as enshrined in the Universal Declaration of Human Rights. They emphasize that this right underscores the importance of \textit{ equal access to the benefits of scientific progress for all}, a principle directly supported by our proposal for Knowledge Units. The UN Special Rapporteur on Cultural Rights further reinforces this, advocating for the expansion of copyright exceptions to broaden access to scientific knowledge as a crucial component of the right to science and culture \cite{scienceright}. 

However, current intellectual property regimes often create ``patently unfair" barriers to this knowledge, preventing innovation and access, especially in areas critical to human rights, as Hale compellingly argues \cite{patentlyunfair}. Finding a solution requires carefully balancing the imperative of open access with the legitimate rights of authors. As Austin and Ginsburg remind us, authors' rights are also human rights, necessitating robust protection \cite{authorhumanrights}. Shareable knowledge entities like Knowledge Units offer a potential mechanism to achieve this delicate balance in the scientific domain, enabling wider dissemination of research findings while respecting authors' fundamental rights.

\vspace{-0.2cm}
\subsection{Impact Across Sectors}

\textbf{Researchers:} Collaboration across different fields becomes easier when knowledge is shared openly. For instance, combining machine learning with biology or applying quantum principles to cryptography can lead to important breakthroughs. Removing copyright restrictions allows researchers to freely use data and methods, speeding up discoveries while respecting original contributions.

\textbf{Practitioners:} Professionals, especially in healthcare, benefit from immediate access to the latest research. Quick access to newer insights on the effectiveness of drugs, and alternative treatments speeds up adoption and awareness, potentially saving lives. Additionally, open knowledge helps developing countries gain access to health innovations.

\textbf{Education:} Education becomes more accessible when teachers use the latest research to create up-to-date curricula without prohibitive costs. Students can access high-quality research materials and use LM assistance to better understand complex topics, enhancing their learning experience and making high-quality education more accessible.

\textbf{Public Trust:} When information is transparent and accessible, the public can better understand and trust decision-making processes. Open access to government policies and industry practices allows people to review and verify information, helping to reduce misinformation. This transparency encourages critical thinking and builds trust in scientific and governmental institutions.

Overall, making scientific knowledge accessible supports global fairness. By viewing knowledge as a common resource rather than a product to be sold, we can speed up innovation, encourage critical thinking, and empower communities to address important challenges.

\vspace{-0.2cm}
\section{Open Problems}
\vspace{-0.1cm}

Moving forward, we identify key research directions to further exploit the potential of converting original texts into shareable knowledge entities such as demonstrated by the conversion into Knowledge Units in this work:

\textbf{1. Enhancing Factual Accuracy and Reliability:}  Refining KUs through cross-referencing with source texts and incorporating community-driven correction mechanisms, similar to Wikipedia, can minimize hallucinations and ensure the long-term accuracy of knowledge-based datasets at scale.

\textbf{2. Developing Applications for Education and Research:}  Using KU-based conversion for datasets to be employed in practical tools, such as search interfaces and learning platforms, can ensure rapid dissemination of any new knowledge into shareable downstream resources, significantly improving the accessibility, spread, and impact of KUs.

\textbf{3. Establishing Standards for Knowledge Interoperability and Reuse:}  Future research should focus on defining standardized formats for entities like KU and knowledge graph layouts \citep{lenat1990cyc}. These standards are essential to unlock seamless interoperability, facilitate reuse across diverse platforms, and foster a vibrant ecosystem of open scientific knowledge. 

\textbf{4. Interconnecting Shareable Knowledge for Scientific Workflow Assistance and Automation:} There might be further potential in constructing a semantic web that interconnects publicly shared knowledge, together with mechanisms that continually update and validate all shareable knowledge units. This can be starting point for a platform that uses all collected knowledge to assist scientific workflows, for instance by feeding such a semantic web into recently developed reasoning models equipped with retrieval augmented generation. Such assistance could assemble knowledge across multiple scientific papers, guiding scientists more efficiently through vast research landscapes. Given further progress in model capabilities, validation, self-repair and evolving new knowledge from already existing vast collection in the semantic web can lead to automation of scientific discovery, assuming that knowledge data in the semantic web can be freely shared.

We open-source our code and encourage collaboration to improve extraction pipelines, enhance Knowledge Unit capabilities, and expand coverage to additional fields.

\vspace{-0.2cm}
\section{Conclusion}
\vspace{-0.1cm}

In this paper, we highlight the potential of systematically separating factual scientific knowledge from protected artistic or stylistic expression. By representing scientific insights as structured facts and relationships, prototypes like Knowledge Units (KUs) offer a pathway to broaden access to scientific knowledge without infringing copyright, aligning with legal principles like German \S 24(1) UrhG and U.S. fair use standards. Extensive testing across a range of domains and models shows evidence that Knowledge Units (KUs) can feasibly retain core information. These findings offer a promising way forward for openly disseminating scientific information while respecting copyright constraints.

\section*{Author Contributions}

Christoph conceived the project and led organization. Christoph and Gollam led all the experiments. Nick and Huu led the legal aspects. Tawsif led the data collection. Ameya and Andreas led the manuscript writing. Ludwig, Sören, Robert, Jenia and Matthias provided feedback. advice and scientific supervision throughout the project. 

\section*{Acknowledgements}

The authors would like to thank (in alphabetical order): Sebastian Dziadzio, Kristof Meding, Tea Mustać, Shantanu Prabhat for insightful feedback and suggestions.  We thank the Grass Foundation and Andrej Radonjic, with special thanks to Andrii Prolorenzo of Wynd Labs for help in scaling up data collection. GR and SA acknowledge financial support by the German Research Foundation (DFG) for the NFDI4DataScience Initiative (project number 460234259). AP and MB acknowledge financial support by the Federal Ministry of Education and Research (BMBF), FKZ: 011524085B and Open Philanthropy Foundation funded by the Good Ventures Foundation. AH acknowledges financial support by the Federal Ministry of Education and Research (BMBF), FKZ: 01IS24079A and the Carl Zeiss Foundation through the project "Certification and Foundations of Safe ML Systems" as well as the support from the International Max Planck Research School for Intelligent Systems (IMPRS-IS). JJ acknowledges funding by the Federal Ministry of Education and Research of Germany (BMBF) under grant no. 01IS22094B (WestAI - AI Service Center West), under grant no. 01IS24085C (OPENHAFM) and under the grant 16HPC117K (MINERVA), as well as co-funding by EU from EuroHPC Joint Undertaking programm under grant no. 101182737 (MINERVA) and from Digital Europe Programme under grant no. 101195233 (openEuroLLM)

%% file: supplementary/appendix.tex
\onecolumn
\section{Appendix}
\subsection{Example Knowledge Unit}
\label{sec:example_ku}
The following shows the first of six KUs of "Sparsity-certifying Graph Decompositions" \cite{streinu2009sparsity}

\textbf{Title:} Sparsity-certifying Graph Decompositions\\
\textbf{Authors:} Ileana Streinu; Louis Theran\\
\textbf{Genre:} Academic Journal, Mathematics, Computer Science\\
\textbf{Style:} The writing style is formal and highly technical, employing specialized terminology from graph theory and matroid theory.  The tone is objective and expository, characteristic of a research paper presenting new algorithms and theoretical results. The text is dense with mathematical notation and\\
\subsubsection*{Knowledge Unit 1 of 6}
\textbf{Context.} The provided context is empty.  Therefore, this summary will describe the text excerpt. The paper introduces a novel algorithm, the (k, $\lambda$)-pebble game with colors, to characterize (k, $\lambda$)-sparse graphs.  This algorithm offers solutions to problems involving tree decompositions of graphs, a topic relevant to rigidity theory. The research builds upon and extends prior work in the field, providing a new proof of the Tutte-Nash-Williams arboricity characterization.

\lstset{
    basicstyle=\ttfamily\small,
    columns=fullflexible,
    keepspaces=true,
    frame=single,
    backgroundcolor=\color{gray!10},
    keywordstyle=\color{blue},
    showstringspaces=false,
    breaklines=true,        
    postbreak=\mbox{\textcolor{red}{$\hookrightarrow$}\space},  
}

\begin{lstlisting}[language=TeX]
'Ileana Streinu': {
  'relations': {
      'authored': 'Sparsity-certifying Graph Decompositions',
      'affiliated_with': 'Smith College',
      'email': 'streinu@cs.smith.edu'
  },
  'attributes': {
      'department': 'Computer Science'
  }
},
'Louis Theran': {
  'relations': {
      'authored': 'Sparsity-certifying Graph Decompositions',
      'affiliated_with': 'University of Massachusetts Amherst',
      'email': 'theran@cs.umass.edu'
  },
  'attributes': {
      'department': 'Computer Science'
  }
},
'Sparsity-certifying Graph Decompositions': {
  'relations': {
      'authors': ['Ileana Streinu', 'Louis Theran'],
      'introduces': '(k, $\lambda$)-pebble game with colors',
      'characterizes': '(k, $\lambda$)-sparse graphs',
      'provides_solutions_for': 'Tree decompositions of graphs',
      'extends_work_of': ['Lee and Streinu', 'Gabow', 'Gabow and Westermann', 'Hendrickson'],
      'proves': 'Tutte-Nash-Williams characterization of arboricity'
  },
  'attributes': {
      'type': 'Academic Journal, Mathematics, Computer Science',
      'topic': 'Graph decompositions',
      'focus': '(k, $\lambda$)-sparse graphs'
  }
},
'(k, $\lambda$)-pebble game with colors': {
  'relations': {
      'introduced_in': 'Sparsity-certifying Graph Decompositions',
      'generalizes': 'Previous results of Lee and Streinu'
  },
  'attributes': {
      'type': 'Algorithm'
  }
},
'(k, $\lambda$)-sparse graphs': {
  'relations': {
      'characterized_by': '(k, $\lambda$)-pebble game with colors',
      'definition': 'No subset of n vertices spans more than $k n - \lambda$ edges'
  },
  'attributes': {
      'range': '$k \geq \lambda \geq 2k-1$ (upper range), $0 \geq \lambda \geq k$ (lower range)'
  }
},
'(k, $\lambda$)-tight graphs': {
  'relations': {
      'are_a_type_of': '(k, $\lambda$)-sparse graphs'
  },
  'attributes': {
      'edge_count': '$k n - $\lambda$'
  }
},
'Tree decompositions of graphs': {
  'relations': {
      'addressed_by': 'Sparsity-certifying Graph Decompositions'
  },
  'attributes': {
      'relevance': 'Rigidity theory'
  }
},
'Tutte-Nash-Williams characterization of arboricity': {
  'relations': {
      'proven_by': 'Sparsity-certifying Graph Decompositions'
  },
  'attributes': {
      'type': 'Theorem'
  }
},
'Decomposition (certifying sparsity)': {
  'relations': {
      'based_on': '(k, $\lambda$)-pebble game with colors',
      'presented_in': 'Sparsity-certifying Graph Decompositions'
  },
  'attributes': {
      'property': 'Sparse graphs and graphs admitting the decomposition coincide'
  }
},
'Algorithms (efficient)': {
  'relations': {
      'presented_in': 'Sparsity-certifying Graph Decompositions',
      'apply_to': 'Upper range of $\lambda$'
  },
  'attributes': {
      'purpose': 'Finding decompositions that certify sparsity'
  }
},
'Previous work': {
  'relations': {
      'referenced_by': 'Sparsity-certifying Graph Decompositions',
      'authors': ['Gabow', 'Gabow and Westermann', 'Hendrickson']
  }
}
\end{lstlisting}

\newpage

\section{Example Multiple-Choice Questions}

\subsection{Abstract-Level Analysis}

\begin{table}[h!]
    \centering
    \caption{Abstract \cite{virnau2006intricate} and generated multiple-choice questions using the Gemini Pro 1.5 002 model. This is a paper from the Medical domain. Correct answers indicated by "*".}
    \label{tab:abstract_medical_mcqs}
    \begin{tabular}{p{0.5\textwidth}|p{0.5\textwidth}}
    \hline
    \textbf{Abstract} & \textbf{Questions} \\
    \hline
    \multirow{3}{0.5\textwidth}{A number of recently discovered protein structures incorporate a rather unexpected structural feature: a knot in the polypeptide backbone. These knots are extremely rare, but their occurrence is likely connected to protein function in as yet unexplored fashion. Our analysis of the complete Protein Data Bank reveals several new knots which, along with previously discovered ones, can shed light on such connections. In particular, we identify the most complex knot discovered to date in human ubiquitin hydrolase, and suggest that its entangled topology protects it against unfolding and degradation by the proteasome. Knots in proteins are typically preserved across species and sometimes even across kingdoms. However, we also identify a knot which only appears in some transcarbamylases while being absent in homologous proteins of similar structure. The emergence of the knot is accompanied by a shift in the enzymatic function of the protein. We suggest that the simple insertion of a short DNA fragment into the gene may suffice to turn an unknotted into a knotted structure in this protein.
} & \begin{minipage}[t]{\linewidth}
        \textit{Which of the following best describes the relationship between protein knots and their function?}\\
        A) Knots invariably enhance protein stability and resistance to degradation.\\
        B) The presence of knots always indicates a shared evolutionary ancestry between proteins.\\
        C*) Knots can influence protein function, but the exact mechanisms are still being investigated.\\
        D) Knot formation is a random event with negligible impact on protein activity.
        
    \end{minipage} \\
    & \vspace{0.2cm}\begin{minipage}[t]{\linewidth}
        \textit{The study of protein knots in ubiquitin hydrolase suggests that:}\\ A) The knot's complexity hinders its interaction with the proteasome.\\
        B*) The knot protects the protein from denaturation and breakdown by the proteasome.\\
        C) The knot facilitates the protein's interaction with other ubiquitin-related proteins.\\
        D) The knot is a result of post-translational modifications.
    \end{minipage} \\
    & \vspace{0.2cm}\begin{minipage}[t]{\linewidth}
        \textit{The example of transcarbamylases illustrates that:} \\
        A) Knots in proteins are always conserved across species and kingdoms. \\
        B*) The emergence of a knot can be linked to a change in the protein's enzymatic activity. \\
        C) Knot formation requires significant alterations to the protein's primary structure. \\
        D) Unknotted proteins are inherently less efficient than their knotted counterparts.
    \end{minipage} \\
    \hline
    \end{tabular}
\end{table}

\newpage

\begin{table}[h!]
    \centering
    \caption{Abstract \cite{DBLP:journals/gc/StreinuT09} and generated multiple-choice questions using the Gemini Pro 1.5 002 model. This is a paper from the Mathematics domain. Correct answers indicated by "*".}
    \label{tab:abstract_math_mcqs}
    \begin{tabular}{p{0.5\textwidth}|p{0.5\textwidth}}
    \hline
    \textbf{Abstract} & \textbf{Questions} \\
    \midrule
    \multirow{3}{0.5\textwidth}{We describe a new algorithm, the $(k,\ell)$-pebble game with colors, and use it to obtain a characterization of the family of $(k,\ell)$-sparse graphs and algorithmic solutions to a family of problems concerning tree decompositions of graphs. Special instances of sparse graphs appear in rigidity theory and have received increased attention in recent years. In particular, our colored pebbles generalize and strengthen the previous results of Lee and Streinu and give a new proof of the Tutte-Nash-Williams characterization of arboricity. We also present a new decomposition that certifies sparsity based on the $(k,\ell)$-pebble game with colors. Our work also exposes connections between pebble game algorithms and previous sparse graph algorithms by Gabow, Gabow and Westermann and Hendrickson.
} & \begin{minipage}[t]{\linewidth}
        \textit{Which of the following best describes the relationship between the $(k,\ell)$-pebble game with colors and the Tutte-Nash-Williams characterization of arboricity, according to the text?}\\
        A) The pebble game provides a counterexample to the Tutte-Nash-Williams characterization.\\
        B*) The pebble game offers new proof and strengthens previous results related to the Tutte-Nash-Williams characterization. \\
        C) The Tutte-Nash-Williams characterization is a specific instance of the $(k,\ell)$-pebble game with colors. \\
        D) The pebble game and the Tutte-Nash-Williams characterization address unrelated graph properties.
    \end{minipage} \\
    
    & \vspace{0.2cm}\begin{minipage}[t]{\linewidth}
        \textit{The described algorithm connects pebble game algorithms with prior sparse graph algorithms by which of the following researchers?}\\ 
        A) Dijkstra and Kruskal \\
        B) Prim and Tarjan \\
        C*) Gabow, Gabow and Westermann, and Hendrickson \\
        D) Ford and Fulkerson
    \end{minipage} \\
    & \vspace{0.2cm}\begin{minipage}[t]{\linewidth}
        \textit{The "new decomposition" mentioned in the text certifies sparsity based on which of the following?} \\
        A) The chromatic number of the graph \\
        B*) The $(k,\ell)$-pebble game with colors \\
        C) The maximum flow through the graph \\
        D) The minimum spanning tree of the graph
    \end{minipage} \\
    \midrule
    \end{tabular}
\end{table}

\newpage

\begin{table}[h!]
    \centering
    \caption{Abstract \cite{DBLP:journals/corr/abs-0704-0671} and generated multiple-choice questions using the Gemini Pro 1.5 002 model. This is a paper from the Computer Science domain. Correct answers indicated by "*".}
    \label{tab:abstract_cs_mcqs}
    \begin{tabular}{p{0.5\textwidth}|p{0.5\textwidth}}
    \hline
    \textbf{Abstract} & \textbf{Questions} \\
    \midrule
    \multirow{3}{0.5\textwidth}{The problem of statistical learning is to construct a predictor of a random variable $Y$ as a function of a related random variable $X$ on the basis of an i.i.d. training sample from the joint distribution of $(X,Y)$. Allowable predictors are drawn from some specified class, and the goal is to approach asymptotically the performance (expected loss) of the best predictor in the class. We consider the setting in which one has perfect observation of the $X$-part of the sample, while the $Y$-part has to be communicated at some finite bit rate. The encoding of the $Y$-values is allowed to depend on the $X$-values. Under suitable regularity conditions on the admissible predictors, the underlying family of probability distributions and the loss function, we give an information-theoretic characterization of achievable predictor performance in terms of conditional distortion-rate functions. The ideas are illustrated on the example of nonparametric regression in Gaussian noise.

} & \begin{minipage}[t]{\linewidth}
        \textit{What is the primary challenge addressed in the described statistical learning problem when the Y-part of the sample is communicated at a finite bit rate?}\\
        A) Reconstructing the joint distribution of (X,Y) with minimal error.\\
        B) Minimizing the computational complexity of encoding the Y-values.\\
        C*) Balancing predictor performance against the constraints imposed by the limited bit rate for Y.\\
        D) Determining the optimal bit rate allocation between X and Y for achieving a desired prediction accuracy.
    \end{minipage} \\
    
    & \vspace{0.2cm}\begin{minipage}[t]{\linewidth}
        \textit{Under what circumstances does the information-theoretic characterization of achievable predictor performance hold, in terms of conditional distortion-rate functions?}\\ A) When the loss function is convex and the admissible predictors are drawn from a parametric class. \\
        B) When the training sample is drawn from a non-i.i.d. distribution and the predictors are nonparametric. \\
        C*) When suitable regularity conditions are met on admissible predictors, the underlying probability distributions, and the loss function. \\
        D) When the X-part of the sample is partially observed and the Y-part is communicated at an infinite bit rate.
    \end{minipage} \\
    & \vspace{0.2cm}\begin{minipage}[t]{\linewidth}
        \textit{How is the concept of conditional distortion-rate functions related to predictor performance in the given scenario?} \\
        A) It quantifies the trade-off between the complexity of the predictor class and the achievable prediction accuracy. \\
        B) It establishes a lower bound on the expected loss of any predictor given the finite bit rate constraint on Y. \\
        C*) It characterizes the achievable predictor performance by quantifying the trade-off between the distortion in representing Y and the bit rate used. \\
        D) It provides a method for selecting the optimal predictor from the admissible class based on the observed X-values.
    \end{minipage} \\
    \midrule
    \end{tabular}
\end{table}

\newpage

\begin{table}[h!]
    \centering
    \caption{Abstract \cite{pan2007evolution} and generated multiple-choice questions using the Gemini Pro 1.5 002 model. This is a paper from the Physics domain. Correct answers indicated by "*".}
    \label{tab:abstract_phy_mcqs}
    \begin{tabular}{p{0.5\textwidth}|p{0.5\textwidth}}
    \hline
    \textbf{Abstract} & \textbf{Questions} \\
    \midrule
    \multirow{3}{0.5\textwidth}{
  The evolution of the Earth-Moon system is described by the dark matter field fluid model proposed in the Meeting of Division of Particle and Field 2004, American Physical Society. The current behavior of the Earth-Moon system agrees with this model very well and the general pattern of the evolution of the Moon-Earth system described by this model agrees with geological and fossil evidence. The closest distance of the Moon to Earth was about 259000 km at 4.5 billion years ago, which is far beyond the Roche's limit. The result suggests that the tidal friction may not be the primary cause for the evolution of the Earth-Moon system. The average dark matter field fluid constant derived from Earth-Moon system data is $4.39 x 10^(-22) s^(-1)m^(-1)$. This model predicts that the Mars's rotation is also slowing with the angular acceleration rate about $-4.38 x 10^(-22) rad s^(-2)$.
    
} & \begin{minipage}[t]{\linewidth}
        \textit{What is the primary implication of the dark matter field fluid model's agreement with the current Earth-Moon system behavior and geological evidence?}\\
        A) Tidal forces are the primary driver of the Earth-Moon system's evolution.\\
        B) The Moon originated from a collision between Earth and a Mars-sized object.\\
        C) The Moon's closest approach to Earth was within the Roche limit.\\
        D*) The tidal friction may not be the primary influence on the Earth-Moon system's evolution.
    \end{minipage} \\
    
    & \vspace{0.2cm}\begin{minipage}[t]{\linewidth}
        \textit{According to the dark matter field fluid model, what was the approximate distance between the Earth and the Moon 4.5 billion years ago?}\\ 
        A) 125,000 km\\
        B*) 259,000 km\\
        C) 384,400 km\\
        D) 450,000 km
    \end{minipage} \\
    & \vspace{0.2cm}\begin{minipage}[t]{\linewidth}
        \textit{The passage mentions a dark matter field fluid constant derived from Earth-Moon system data.  Which of the following best describes the significance of this constant in relation to Mars?} \\
        A) It predicts the rate of decrease in Mars's orbital velocity.\\
        B) It calculates the rate at which Mars's magnetic field is decaying.\\
        C*) It predicts the angular acceleration rate of Mars's rotation.\\
        D) It estimates the rate of expansion of Mars's crust due to internal heating.
    \end{minipage} \\
    \midrule
    \end{tabular}
\end{table}

\newpage

\subsection{Full-Paper Analysis}

\begin{table}[h!]
    \centering
    \caption{Eight of ten generated multiple-choice questions by the Gemini Pro 1.5 002 model from \cite{chabot2014interrelationships}. This is a paper from the Medical domain. Correct answers indicated by "*".}
    \label{tab:fullpaper_medical_mcqs}
    \begin{tabular}{p{0.5\textwidth}|p{0.5\textwidth}}
    \hline
    \multicolumn{2}{c}{\textbf{Questions}} \\
    \midrule
    \begin{minipage}[t]{\linewidth}
    \textit{What is the primary mechanism by which ghrelin stimulates growth hormone (GH) release?}\\
    A) Direct activation of the growth hormone releasing hormone receptor (GHRHr)\\
    B) Stimulation of somatostatin release from the hypothalamus\\
    C*) Activation of the ghrelin receptor (GHSR), specifically the GHSR1a subtype\\
    D) Modulation of opioid peptide activity in the pituitary
    \end{minipage} & \begin{minipage}[t]{\linewidth}
    \textit{Which statement best describes the relationship between ghrelin and insulin sensitivity in healthy individuals after acute administration?}\\
    A) Ghrelin enhances insulin sensitivity, leading to increased glucose uptake.\\
    B) Ghrelin has no significant effect on insulin sensitivity.\\
    C*) Ghrelin impairs insulin sensitivity, potentially through direct effects on the liver.\\
    D) Ghrelin's effect on insulin sensitivity is entirely dependent on GH secretion.
    \end{minipage} \\
    \midrule
    \begin{minipage}[t]{\linewidth}
    \textit{What is the primary site of ghrelin production in the body?}\\
    A) Hypothalamus\\
    B) Pituitary gland\\
    C*) Stomach\\
    D) Small intestine
    \end{minipage} & \begin{minipage}[t]{\linewidth}
    \textit{How does unacylated ghrelin (UAG) affect the metabolic actions of acylated ghrelin (AG)?}\\
    A) UAG amplifies the hyperglycemic and hyperinsulinemic effects of AG.\\
    B) UAG has no impact on the metabolic effects of AG.\\
    C*) UAG counteracts the hyperglycemic and hyperinsulinemic effects of AG.\\
    D) UAG mimics the effects of AG on GH secretion but not on glucose metabolism.
    \end{minipage} \\
    \midrule
    \begin{minipage}[t]{\linewidth}
    \textit{What is the role of GOAT in ghrelin processing?}\\
    A) Cleavage of the ghrelin precursor into its active form\\
    B*) Octanoylation of the serine-3 residue of ghrelin\\
    C) Deacylation of acylated ghrelin to form UAG\\
    D) Binding and transport of ghrelin in the bloodstream
    \end{minipage} & \begin{minipage}[t]{\linewidth}
    \textit{What is the effect of long-term ghrelin treatment on plasma glucose and insulin levels?}\\
    A) Consistently decreases both glucose and insulin levels\\
    B*) Increases glucose levels, while insulin levels remain unchanged or increase\\
    C) Decreases glucose levels, while insulin levels increase\\
    D) Has no consistent effect on either glucose or insulin levels
    
    \end{minipage} \\
    \midrule
    \begin{minipage}[t]{\linewidth}
    \textit{What role does the vagus nerve play in ghrelin's effects on insulin secretion?}\\
    A) Ghrelin stimulates the vagus nerve to enhance insulin secretion.\\
    B*) Ghrelin inhibits the vagus nerve to suppress insulin secretion, particularly through the hepatic branch.\\
    C) Ghrelin's effects on insulin secretion are independent of vagal activity.\\
    D) Ghrelin acts synergistically with vagal stimulation to increase insulin secretion.
    \end{minipage} & \begin{minipage}[t]{\linewidth}

    \textit{Which of the following best describes the effect of ghrelin on glucose-stimulated insulin secretion (GSIS) in isolated pancreatic islets and cell lines?}
    A) Ghrelin consistently enhances GSIS.\\
    B) Ghrelin consistently inhibits GSIS.\\
    C) Ghrelin has no effect on GSIS.\\
    D*) Ghrelin's effect on GSIS is complex and may depend on factors like glucose concentration and ghrelin dose.
    \end{minipage} \\
    \midrule
    \end{tabular}
\end{table}

\begin{table}[h!]
    \centering
    \caption{Eight of ten generated multiple-choice questions by the Gemini Pro 1.5 002 model from \cite{cipolloni2023entanglement}. This is a paper from the Physics domain. Correct answers indicated by "*".}
    \label{tab:fullpaper_phy_mcqs}
    \begin{tabular}{p{0.5\textwidth}|p{0.5\textwidth}}
    \hline
    \multicolumn{2}{c}{\textbf{Questions}} \\
    \midrule
    \begin{minipage}[t]{\linewidth}
    \textit{What is the primary focus of the paper discussed in the text?}\\
    A) Developing a new definition of entanglement entropy in gauge theories.\\
    B) Exploring the entanglement structure of strongly coupled Yang-Mills theories.\\
    C*) Utilizing recent technical advancements to understand ground state entanglement in weakly coupled Yang-Mills theories.\\
    D) Comparing different approaches to calculating entanglement entropy in gauge theories and establishing their equivalence.
    \end{minipage} & \begin{minipage}[t]{\linewidth}
    \textit{What is the main difficulty in defining entanglement entropy in gauge theories?}\\
    A) The non-Abelian nature of the gauge group makes it challenging to define subsystems.\\
    B) Gauge invariance introduces nonlocality at the UV scale, making subsystem definition difficult.\\
    C) The presence of both electric and magnetic terms in the Hamiltonian complicates the calculation.\\
    D*) The lack of a clear separation between physical and unphysical degrees of freedom makes it hard to define a reduced density operator.
    \end{minipage} \\
    \midrule
    \begin{minipage}[t]{\linewidth}
    \textit{Which approach does the paper primarily follow to define entanglement entropy?}\\
    A) Embedding the physical Hilbert space into a larger direct product space.\\
    B*) Using the replica trick and Euclidean path integral methods.\\
    C) Employing the Ryu-Takayanagi prescription in the holographic dual.\\
    D) Constructing a gauge-invariant density operator within a subalgebra of observables.
    \end{minipage} & \begin{minipage}[t]{\linewidth}
    \textit{How does the paper address the issue of the physical Hilbert space not admitting a direct product decomposition?}\\
    A) It introduces a new type of gauge-invariant operator that allows for a direct product decomposition.\\
    B) It utilizes a gauge-fixing procedure that eliminates the nonlocal effects of gauge invariance.\\
    C*) It works with an extended basis of gauge-variant states and accounts for the entropy contribution from splitting flux lines.\\
    D) It restricts the algebra of observables to a subalgebra that does admit a direct product decomposition.
    \end{minipage} \\
    \midrule
    \begin{minipage}[t]{\linewidth}
    \textit{What is the significance of the ubiquitous term identified in the entanglement entropy of Yang-Mills theories?}\\
    A) It represents the contribution of edge modes to the entanglement entropy.\\
    B*) It is a universal term that dominates the entanglement entropy in 3+1 dimensions.\\
    C) It arises from the presence of topological defects in the gauge theory.\\
    D) It is a non-universal term that depends on the specific lattice regularization.
    \end{minipage} & \begin{minipage}[t]{\linewidth}
    \textit{How is the Yang-Mills theory related to the principal chiral model in the paper's calculation?}\\
    A) The Yang-Mills theory is dual to the principal chiral model.\\
    B*) The Yang-Mills theory can be expressed as a principal chiral model after gauge fixing to axial gauge.\\
    C) The principal chiral model is used as a toy model to understand the qualitative features of the Yang-Mills theory.\\
    D) The principal chiral model provides a non-perturbative definition of the Yang-Mills theory.
    
    \end{minipage} \\
    \midrule
    \begin{minipage}[t]{\linewidth}
    \textit{What is the role of Nambu-Goldstone bosons in the entanglement entropy calculation?}\\
    A) They represent the gauge degrees of freedom that are fixed in the axial gauge.\\
    B*) Their enhanced entanglement of the softest mode contributes to the logarithmic term in the entropy.\\
    C) They mediate the interactions between the electric and magnetic degrees of freedom.\\
    D) Their zero mode fluctuations determine the Shannon entropy contribution to the entanglement entropy.
    \end{minipage} & \begin{minipage}[t]{\linewidth}

    \textit{What is the connection between the logarithmic term in the entanglement entropy and topological entanglement entropy?}
    A*) The logarithmic term is a generalization of topological entanglement entropy to continuous gauge groups.\\
    B) The logarithmic term is a correction to topological entanglement entropy at weak coupling.\\
    C) The logarithmic term is equivalent to topological entanglement entropy in the planar limit.\\
    D) The logarithmic term is unrelated to topological entanglement entropy.
    \end{minipage} \\
    \midrule
    \end{tabular}
\end{table}

\section{Similarity Overlaps}
\label{sec:appsimilarity}

\begin{figure}[h]
    \centering
    \fbox{\includegraphics[width=\linewidth]{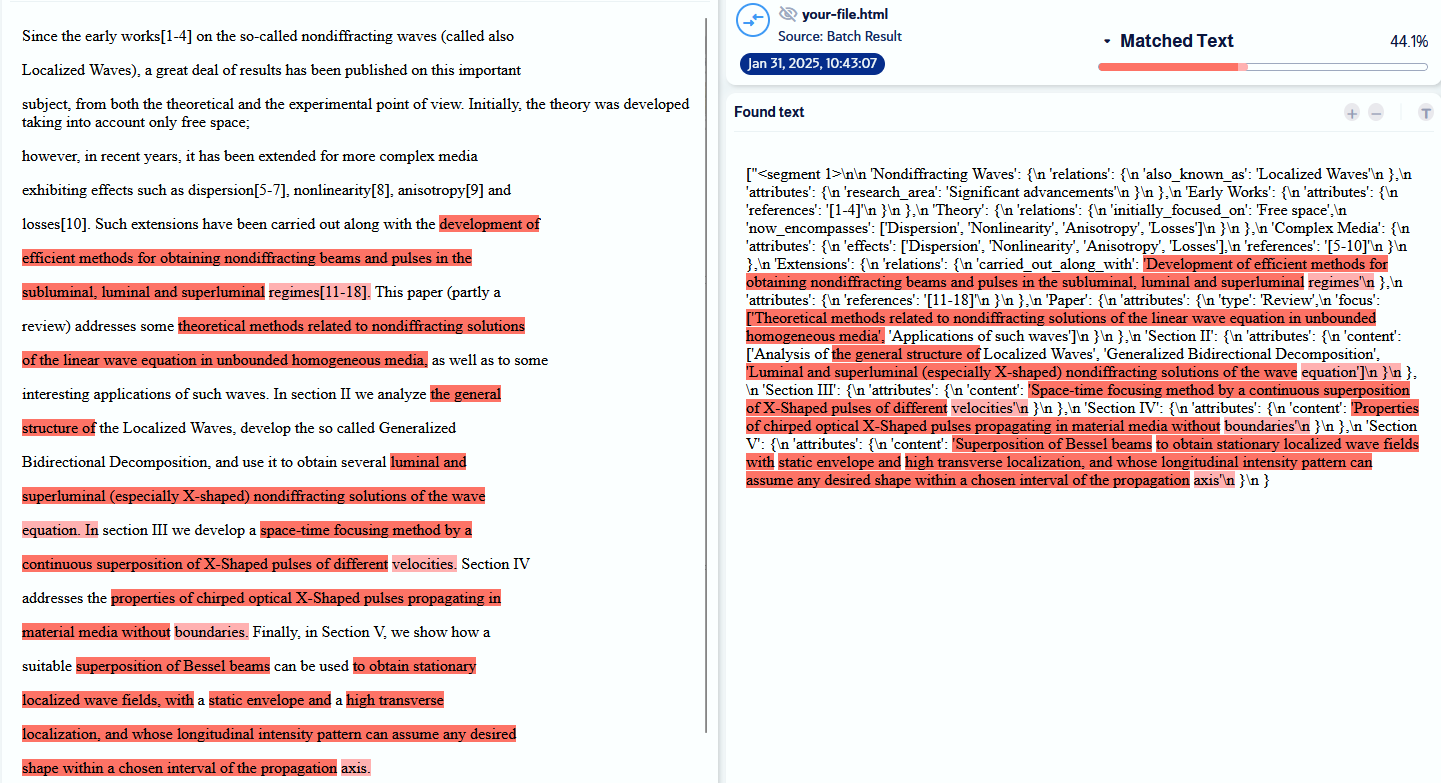}}
    \vspace{1em} 
    \fbox{\includegraphics[width=\linewidth]{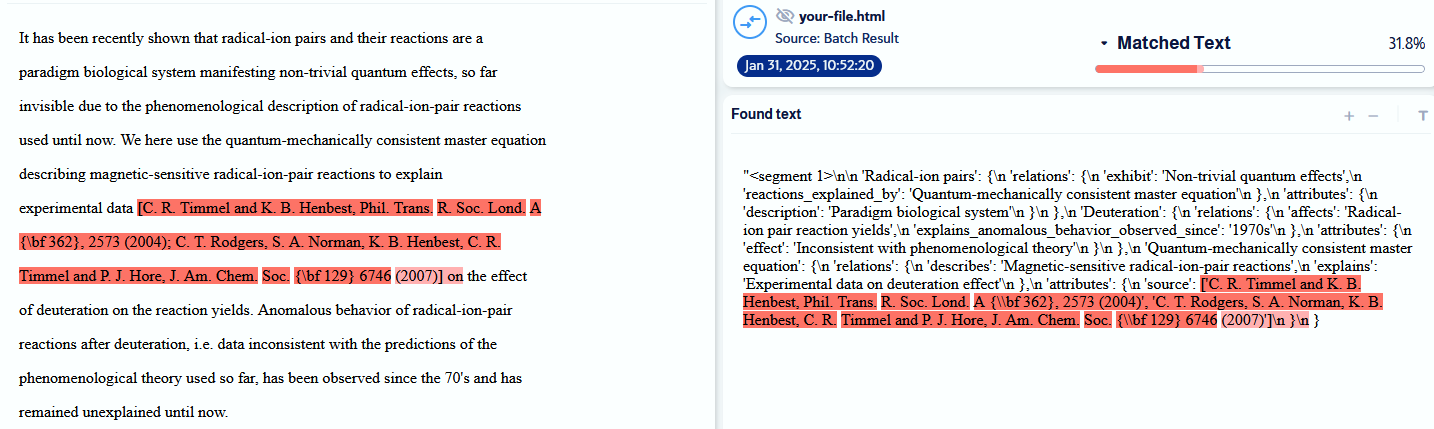}}
    \vspace{1em}
    \fbox{\includegraphics[width=\linewidth]{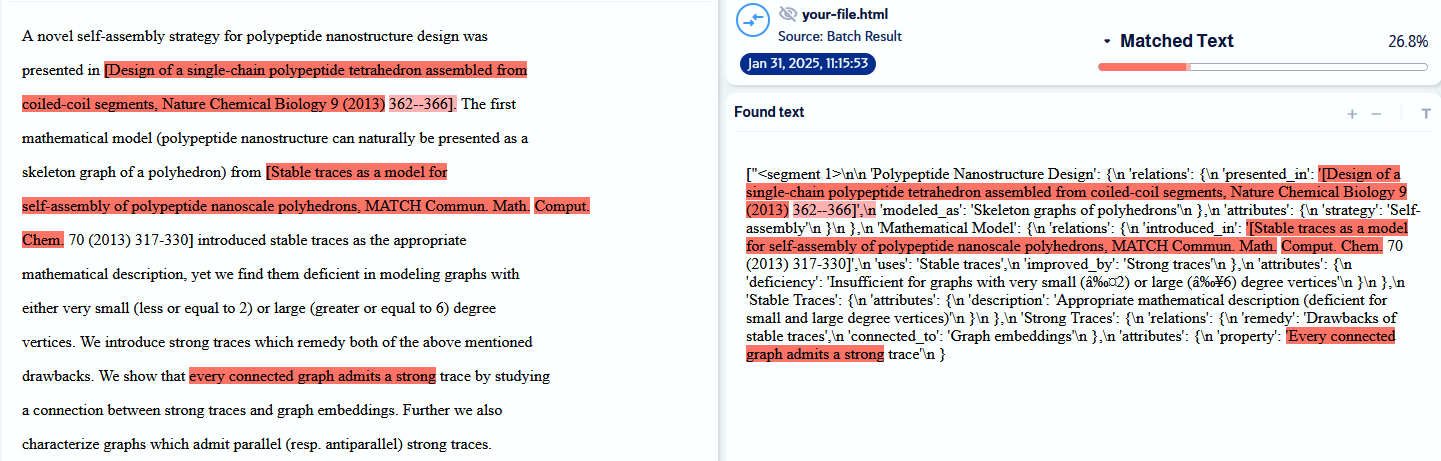}}
    \vspace{1em}
    \caption{\textbf{Similarity Overlaps:} Overlap between Top-3 Most Similar Original Texts and Knowledge Units using an online plagiarism checker \href{https://app.copyleaks.com/text-compare}{tool}.}
    \label{fig:simtop}
\end{figure}

\begin{figure}[h]
    \centering
    \fbox{\includegraphics[width=\linewidth]{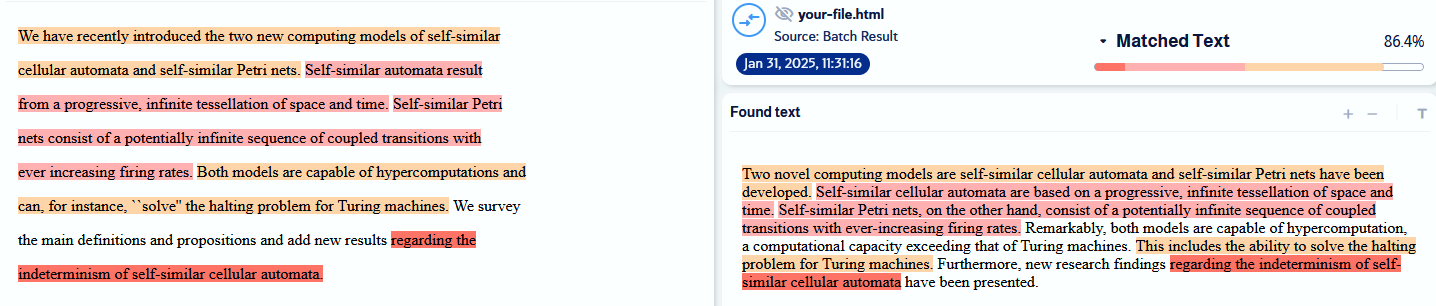}}
    \vspace{1em} 
    \fbox{\includegraphics[width=\linewidth]{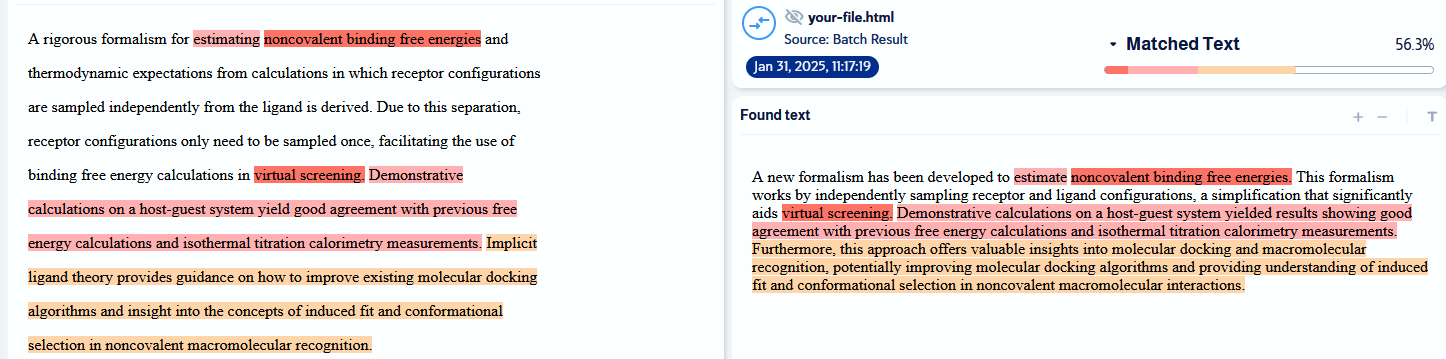}}
    \vspace{1em}
    \fbox{\includegraphics[width=\linewidth]{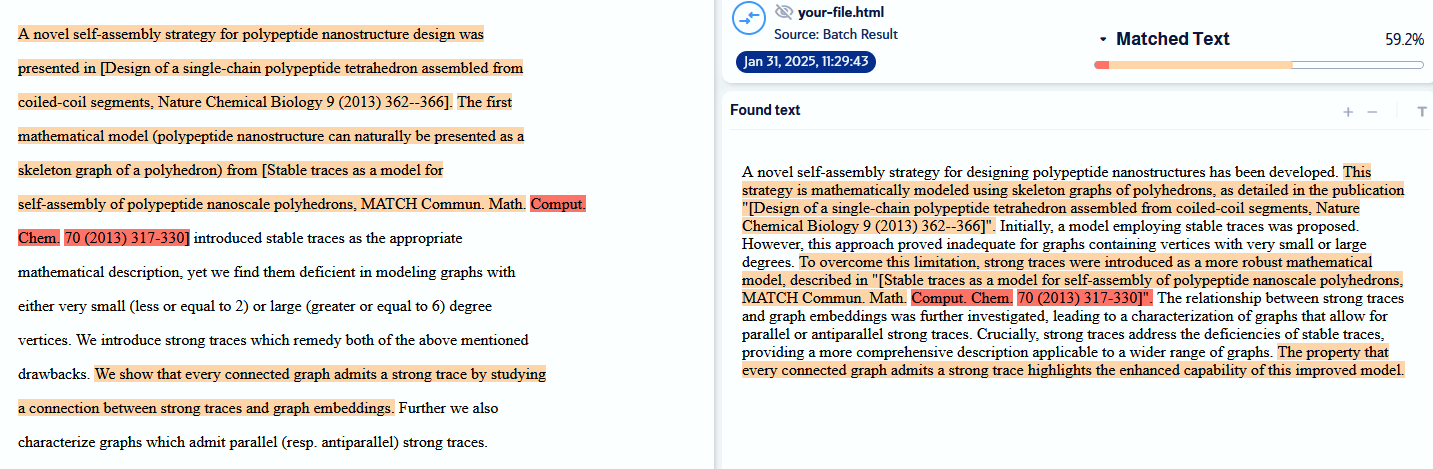}}
    \vspace{1em}
    \caption{\textbf{Similarity Overlaps:} Overlap between Top-3 Most Similar Original Texts and reconstructured text via Knowledge Units using an online plagiarism checker \href{https://app.copyleaks.com/text-compare}{tool}.}
    \label{fig:simtoprecon}
\end{figure}

In Figure \ref{fig:simtop}, we highlight the top similarity overlaps between Original Texts and Knowledge Units, while Figure \ref{fig:simtoprecon} focuses on overlaps between Original Texts and Reconstructed Texts. In both cases, the shared segments predominantly consist of scientific jargon or references to particular issues, illustrating the specialized nature of the content.

\clearpage

\section{Legal Opinion}
\label{sec:legal}

We include, alongside this paper, a legal opinion from a team of lawyers which formed the source material used to derive legal insights about the German law, but not publicly available. This opinion corresponds to \citet{heidrich2023legal} and is provided here for completeness.



\section*{Supplementary material}

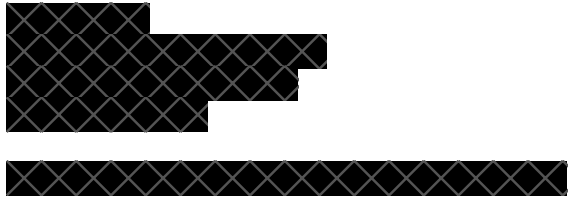

**IT-RECHT
DATENSCHUTZ
COMPLIANCE**

Heidrich Rechtsanwälte PartG mbB

JOERG HEIDRICH (Partner)
Rechtsanwalt, Fachanwalt für IT-Recht

NICK AKINCI, MLE (Partner)
Rechtsanwalt

NIKLAS MÜHLEIS, LL.M. (Partner)
Rechtsanwalt

TIMO BUSCH (angestellt)
Rechtsanwalt

SARAH HENCKE (angestellt)
Rechtsanwältin

Tel.: +49 (511) 374 98 150
Fax: +49 (511) 374 98 151
Mail: kontakt@recht-im-internet.de
Netz: www.recht-im-internet.de

6. Dezember 2023

**Prüfung urheberrechtlicher Fragestellungen**
**Unser Zeichen: 18/23-NA-IT**

Sehr geehrte Damen und Herren,

auftragsgemäß haben wir geprüft, ob die öffentliche Zurverfügungstellung von Wissensgraphen – bestehend aus Informationen aus urheberrechtlich geschützten Werken Dritter – einen Verstoß gegen das deutsche Urheberrecht darstellen kann. Zudem haben wir geprüft, ob das Speichern von Satzfragmenten zur Setzung von sogenannten Ankern urheberrechtliche Relevanz aufweist.

Im Ergebnis dürfte die öffentliche Zurverfügungstellung von Wissensgraphen keine Urheberrechtsverletzung darstellen. Auch die Entnahme von Satzfragmenten dürfte im Ergebnis rechtmäßig sein, wenngleich hier ein gewisses Restrisiko verbleibt. Gern erläutern wir dies näher:

1. **Sachverhalt**

Für die Erstellung der Wissensgraphen werden die ausgewählten urheberrechtlich geschützten Werke zunächst in einer Datenbank gespeichert.

Anschließend werden aus den ausgewählten Texten absatzweise Informationen entnommen

Heidrich Rechtsanwälte
Partnerschaftsgesellschaft mbB
Prinzenstraße 3
30159 Hannover

Sitz der Gesellschaft: Hannover
Registergericht: AG Hannover
Registernummer: PR 201108
Steuernummer: 25/232/49701

Vertretungsberechtigte Partner:
Rechtsanwalt Joerg Heidrich
Rechtsanwalt Nick Akinci
Rechtsanwalt Niklas Mühleis

Honorarkonto:
Commerzbank AG
DE51 2504 0066 0332 6444 00
COBADEFFXXX

Anderkonto:
Commerzbank AG
DE25 2504 0066 0333 4489 01
COBADEFFXXX

und in maschinenlesbarer Form abgespeichert. Dabei werden lediglich die Informationen selbst, nicht jedoch Textteile entnommen. Danach werden die Werke wieder aus der Datenbank gelöscht.

Die gewonnen Informationen werden sodann für die Erstellung von sogenannten Wissensgraphen genutzt. In diesem Rahmen werden die einzelnen Informationen logisch verknüpft und miteinander ins Verhältnis gesetzt (siehe zur Veranschaulichung nachfolgende Abbildung). Dabei werden die Zusammenhänge der Objekte, Orte, Personen und Beziehungen untereinander erfasst.

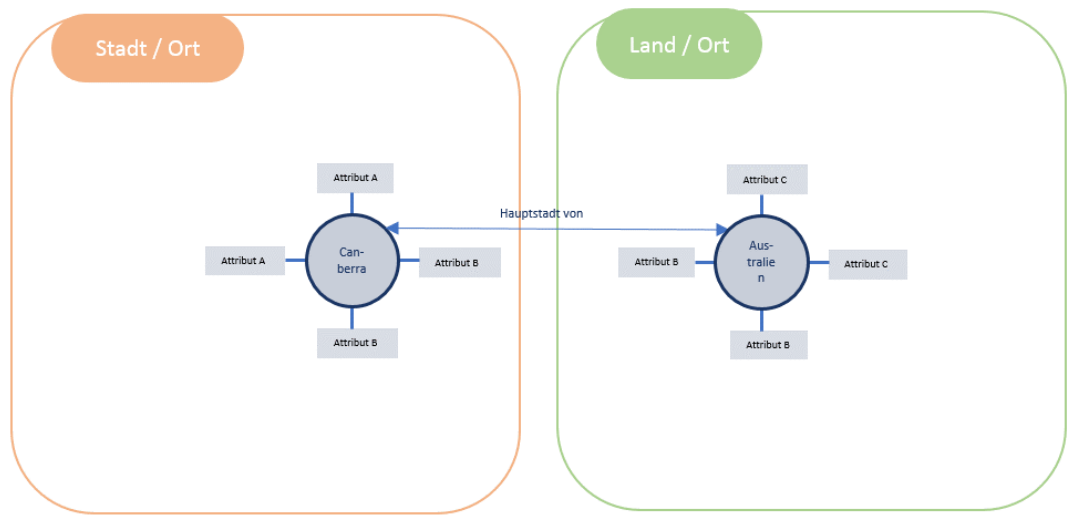

Weiterhin sollen zum Zwecke der Setzung sogenannter Anker, Satzfragmente aus den Texten entnommen werden. Durch eine Abfolge von vier bis fünf Worten, können so bestimmte Stellen im Text auffindbar gemacht werden, um den extrahierten Informationen eine Textstelle zuzuweisen. Die Satzfragmente müssen hierfür dauerhaft gespeichert werden.

## 2. Schutz von Werken

Im Ausgangspunkt erfasst das deutsche Urheberrecht den Schutz von Werken. Nach § 2 Abs. 1 UrhG liegt ein Werk vor, wenn es sich um eine persönliche geistige Schöpfung handelt, die eine bestimmte Schöpfungshöhe erreicht. Diese orientiert sich anhand der Individualität und Originalität eines Werkes.

An die Individualität sind keine allzu hohen Maßstäbe zu stellen. Auch eine Schöpfung mit

geringer Individualität kann die vorausgesetzte Schöpfungshöhe erreichen und folglich Urheberrechtsschutz genießen.

Dem Urheber steht das ausschließliche Recht zur Nutzung zu. Nutzungen im Sinne des Gesetzes sind unter anderem die **Vervielfältigung** und die **öffentliche Zugänglichmachung**.

Dem Urheber steht es frei, anderen die Nutzung seines Werkes zu gestatten und ihnen Nutzungsrechte daran einzuräumen. Liegt eine solche Erlaubnis nicht vor, ist die Nutzung nur im Rahmen der engen gesetzlich definierten Ausnahmen (sogenannte Schranken) des UrhG zulässig. Im Hinblick auf die persönliche geistige Schöpfung ist festzuhalten, dass das deutsche Urheberrecht als Werkschöpfer lediglich Menschen in Betracht zieht, so das von KI generierte Erzeugnisse nach der derzeitigen Rechtslage keinen Urheberrechtsschutz genießen. Lediglich Menschen sind in diesem Sinne dazu befähigt, eine persönliche geistige Schöpfung zu tätigen.

### 3. Speicherung von Werken zu Analysezwecken

Die Speicherung der Texte stellt eine urheberrechtlich relevante Vervielfältigung im Sinne des § 16 UrhG dar, die grundsätzlich nur dem Urheber bzw. Rechteinhaber zusteht. Dabei ist es unerheblich, ob die Speicherung kurz- oder langfristig erfolgt.

Zu prüfen war, ob diese Vervielfältigungen durch die Schrankenbestimmungen zum **Text und Data Mining** gemäß § 44b UrhG oder § 60d UrhG gerechtfertigt sind. Danach sind Vervielfältigungen zum Zwecke des Text und Data Minings ohne Zustimmung des Urhebers zulässig. Text und Data Mining wird in § 44b Abs. 1 UrhG definiert als die automatisierte Analyse von einzelnen oder mehreren digitalen oder digitalisierten Werken, um daraus Informationen insbesondere über Muster, Trends und Korrelationen zu gewinnen. Ausweislich der Gesetzesbegründung ist das Text und Data Mining für das maschinelle Lernen als Basis-Technologie für Künstliche Intelligenz von besonderer Bedeutung.[1]

Vorliegend erfolgen die Vervielfältigungen der fremden Inhalte, um Informationen aus den Werken zu entnehmen und diese Informationen miteinander zu verknüpfen und ins Verhältnis zu setzen. Damit entspricht der Zweck der Vervielfältigungen dem Wortlaut des § 44b UrhG, wonach insbesondere die Gewinnung von Informationen über Korrelationen erfasst ist.

§ 60d UrhG schützt das Text und Data Mining zum Zweck der wissenschaftlichen Forschung und erlaubt neben Vervielfältigungen auch das öffentliche Zugänglichmachen von Vervielfältigungen. Zudem normiert § 60d UrhG im Gegensatz zu § 44b UrhG dem Urheber

---

[1] Bundestagsdrucksache 19/29894

nicht die Möglichkeit einen Nutzungsvorbehalt auszusprechen, der eine Nutzung seiner Werke untersagt.

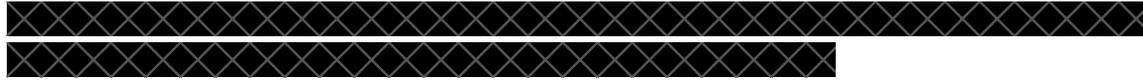

### 4. Erstellung von Wissensgraphen

Zu prüfen war, ob die Entnahme einzelner Informationen zur Aufbereitung als Wissensgraph urheberrechtliche Relevanz hat.

Eine Vervielfältigungshandlung dürfte in diesem Rahmen ausscheiden, da die entnommenen Informationen keinen Schöpfungsgrad aufweisen und der Text nicht kopiert wird.

In Frage kommt jedoch eine sogenannte **unfreie Bearbeitung**. Eine solche wäre gemäß § 23 Abs. 1 S. 1 UrhG wiederrum nur mit Zustimmung des Urhebers rechtmäßig. Eine unfreie Bearbeitung liegt vor, wenn das benutzte Originalwerk auch in der Bearbeitungsfassung erkennbar mit seinen Wesenszügen und Eigenheiten durchscheint.[2] Bejaht wird dies in Bezug auf Abstracts und Zusammenfassungen. Dabei kommt es entscheidend darauf an, ob der eigenschöpferische Gehalt des Ausgangstextes, der insbesondere in den wesentlichen und prägenden Formulierungen und Satzteilen besteht, übernommen wird.[3]

Vor diesem Hintergrund dürfte eine unfreie Bearbeitung vorliegend ausscheiden. Geplant ist lediglich einzelne Informationen, nicht aber Formulierungen oder Satzteile, zu entnehmen. Die Ursprungswerke dürften in den Wissensgraphen gerade nicht mehr zu erkennen sein und somit dürfte der erforderliche Abstand zu den Ursprungswerken gewahrt sein.

Denn die Ausgangstexte zeichnen sich durch die Anordnung der Worte und Sätze als kreative Schöpfung aus. Entnimmt man lediglich isoliert die einzelnen Informationen und setzt diese miteinander ins Verhältnis, so ist das Ursprungswerk regelmäßig nicht mehr zu erkennen. Demnach scheinen in den Wissensgraphen die Wesenszüge und Eigenheiten des Ursprungswerkes in sprachlicher Hinsicht nicht durch. Aus dem Wissensgraphen ist aus der Zusammensetzung der entnommenen Informationen des Ursprungwerkes nicht mehr die Anordnung der Worte und Sätze als kreative Schöpfung zu erkennen.

Zudem liefen die Regelungen der §§ 44b und 60d UrhG leer, wenn die auf Basis dieser Vorschriften rechtmäßig extrahierten Informationen nicht zur Schaffung intelligenter

---

[2] *Loewenheim* in Schricker/Loewenheim, UrhR, 6. Aufl. 2020, § 3 UrhG, Rn. 7.
[3] *Bullinger* in Wandtke/Bullinger, UrhR, 6. Aufl. 2022, § 23 UrhG, Rn. 7.

Datenbanken oder künstlicher Intelligenz verwendet werden könnten.

## 5. Veröffentlichung der Wissensgraphen

Für die Frage, ob die Veröffentlichung der Wissensgraphen gegen das Urheberrecht verstoßen kann, ist entscheidend,

1. ob die Wissensgraphen urheberrechtlichen Schutz genießen und

2. wer als Urheber der Wissensgraphen gilt – sofern es einen solchen gibt.

Die aus den Werken Dritter entnommenen Informationen können in ihrer Zusammensetzung als Wissensgraphen die erforderliche Schöpfungshöhe erreichen. Diese können als Text-, Software- oder Datenbankwerk geschützt sein. Ob ein urheberrechtlicher Schutz entstehen kann, hängt jedoch maßgeblich davon ab, ob die Wissensgraphen durch einen Menschen oder durch eine Software bzw. eine Künstliche Intelligenz erstellt werden. Im letzteren Fall bliebe ein Urheberrechtsschutz aus den unter Punkt 2 genannten Gründen verwehrt.

Da die Wissensgraphen – wie bereits ausgeführt – keine Bearbeitung der analysierten Texte darstellen, können nur die Ersteller der Wissensgraphen (sofern es sich dabei um Menschen handelt) Urheber sein.

Da das Urheberrecht an den Wissensgraphen nur den Erstellern dieser Graphen zustehen kann, verstößt die Veröffentlichung auch nicht gegen die Rechte der Verfasser der Ursprungstexte.

## 6. Entnahme von Satzfragmenten zum Setzen von Ankerpunkten

Zudem haben wir geprüft, ob die Entnahme von Satzfragmenten zum Setzen von Ankerpunkten eine Urheberrechtsverletzung darstellen kann.

Die kopierten Satzfragmente bestehen nach unseren Informationen in der Regel aus wenigen Worten. Diese werden benötigt, um eine bestimmte Stelle aus einem Textwerk zu referenzieren.

Vom Urheberrechtsschutz erfasst sind jedoch nur Worte und Satzfragmente, soweit der wiedergegebene Bestandteil die eigene geistige Schöpfung durch den Urheber zum Ausdruck bringt.[4] Sofern nur bis zu fünf Worte entnommen werden, dürfte es regelmäßig an einer

---

[4] Urteil des EUGH vom 16. Juli 2009 – C-5/08.

geistigen Schöpfung fehlen, sofern die Wortfolge keine besondere Originalität aufweist. Dies dürfte in der Mehrzahl der Fälle so sein.

Neben den Rechten aus § 15 UrhG könnten vorliegend auch die Rechte aus dem sogenannten Presseverleger-Leistungsschutzrecht nach den §§ 87f ff. UrhG berührt sein, sofern die Satzfragmente aus Presseveröffentlichungen im Sinne des § 87f Abs. 1 UrhG entnommen werden. Nicht umfasst davon sind „Periodika, die für wissenschaftliche oder akademische Zwecke verlegt werden".

Nach § 87g UrhG haben Presseverleger das ausschließliche Recht, ihre Presseveröffentlichung im Ganzen oder in Teilen für die Online-Nutzung durch Anbieter von Diensten der Informationsgesellschaft öffentlich zugänglich zu machen und zu vervielfältigen. Davon erfasst sind grundsätzlich auch kleinste Teile von Presseveröffentlichungen. Nach dem Wortlaut von § 87g Abs. 2 Nr. 4 UrhG ist jedoch „die Nutzung einzelner Wörter oder sehr kurzer Auszüge" nicht vom Schutz des Leistungsschutzrechts umfasst.

Im Ergebnis dürfte daher auch das Presseverleger-Leistungsschutzrecht in der Regel nicht berührt sein. Da das Leistungsschutzrecht jedoch tendenziell auch kleinere Teile von Texten schützt, besteht hier im Vergleich zu anderen Texten (die „nur" dem allgemeinen urheberrechtlichen Schutz unterliegen), ein leicht erhöhtes Risiko von Rechtsverletzungen.

Zuletzt kann noch die Frage aufgeworfen werden, ob durch die vielfache Entnahme von Satzfragmenten aus demselben Werk, insgesamt eine Vervielfältigung vorliegen könnte, wenn dadurch der Gesamtinhalt eines Werkes reproduziert wird. Dies erscheint im Hinblick auf literarische Werke wie Romane jedenfalls möglich, da nach der Rechtsprechung des BGH[5] auch Charaktere, deren Umgebungen und Beziehungsgeflechte geschützt sein können. Allerdings ist die Erstellung von Wissensgraphen mit dem dieser Rechtsprechung zugrundeliegenden Fall (in welchem es um die Fortschreibung eines Romans ging) nicht vergleichbar. Bei der Erstellung von Wissensgraphen geht es nicht darum, das Originalwerk fortzuführen und so die Verwertung auszuweiten, sondern um die strukturierte Darstellung von Informationen.

### 7. Zusammenfassung

Zusammenfassend lässt sich Folgendes festhalten:

- Die Speicherung der Texte zur Analyse, dürfte von den Ausnahmeregelungen der §§ 44b und 60d UrhG gedeckt und damit rechtmäßig sein, sofern die Kopien nach der

---

[5] *BGH*, Urteil vom 29.04.1999, Az.: I ZR 65/96.

Analyse umgehend gelöscht werden.

- Die Analyse und Erstellung der Wissensgraphen selbst sowie die Veröffentlichung der Graphen dürfte nicht in die Rechte der Urheber eingreifen und damit ebenfalls rechtmäßig sein.

- Die Entnahme und dauerhafte Speicherung von Satzfragmenten zur Setzung von Ankerpunkten dürfte aller Voraussicht nach ebenfalls nicht rechtsverletzend sein – es verbleibt jedoch ein gewisses Restrisiko.

Sollten hierzu noch Fragen bestehen, stehen wir Ihnen hierfür selbstverständlich gerne zur Verfügung.

Mit freundlichen Grüßen

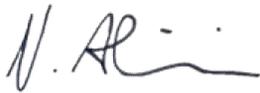

Nick Akinci

Rechtsanwalt

